\pdfoutput=1

\documentclass[11pt]{article}

\usepackage[preprint]{acl}

\usepackage{times}
\usepackage{latexsym}

\usepackage[T1]{fontenc}

\usepackage[utf8]{inputenc}

\usepackage{microtype}

\usepackage{inconsolata}

\usepackage{graphicx}
\usepackage{hyperref}
\usepackage{listings}
\usepackage{longtable}
\usepackage{todonotes}
\usepackage{xspace}
\usepackage[normalem]{ulem}
\usepackage{pifont}
\usepackage{makecell}
\usepackage{amsmath}
\usepackage{dsfont}
\usepackage{soul}
\usepackage{booktabs}
\usepackage{colortbl}
\usepackage{float}

\usepackage{epstopdf}
\DeclareGraphicsExtensions{.pdf,.png}

\newcommand{\user}{\textit{User}\xspace}
\newcommand{\executioncontext}{\textit{Execution Context}\xspace}
\newcommand{\agent}{\textit{Agent}\xspace}
\newcommand{\executionenvironment}{\textit{Execution Environment}\xspace}
\newcommand{\interactiveconsole}{\textit{code.InteractiveConsole}\xspace}
\newcommand{\tool}{\textit{Tool}\xspace}
\newcommand{\tools}{\textit{Tools}\xspace}
\newcommand{\toolsandbox}{\textsc{ToolSandbox}\xspace}
\newcommand{\messagebus}{\textit{Message Bus}\xspace}
\newcommand{\milestones}{\textit{Milestones}\xspace}
\newcommand{\minefields}{\textit{Minefields}\xspace}

\usepackage{bbding}
\newcommand{\xmark}{{\textcolor{red}{\XSolidBrush}}}
\definecolor{ao(english)}{rgb}{0.0, 0.5, 0.0}
\newcommand{\cmark}{{\textcolor{ao(english)}{\CheckmarkBold}}}
\usepackage{tikz}

\newcommand{\halfcheckmark}{{\textcolor{orange}{\tikz\draw[scale=0.4,fill=orange](0,.35) -- (.25,0) -- (1,.7) -- (.25,.15) -- cycle (0.75,0.2) -- (0.77,0.2)  -- (0.6,0.7) -- cycle;}}}

\DeclareMathOperator*{\argmax}{arg\,max}

\definecolor{mycolor}{RGB}{0,0,255}
\newcommand{\scorecolor}[1]{\cellcolor{mycolor!#1}}

\title{\toolsandbox: A Stateful, Conversational, Interactive Evaluation Benchmark for LLM Tool Use Capabilities}

\author{
    Jiarui Lu, Thomas Holleis, Yizhe Zhang, Bernhard Aumayer \\
    {\bf Feng Nan, Felix Bai, Shuang Ma, Shen Ma, Mengyu Li,} \\ 
    {\bf Guoli Yin, Zirui Wang, Ruoming Pang} \\
    Apple \\
    \texttt{\{jiarui\_lu, tholleis, yizhe\_zhang, baumayer} \\
    \texttt{f\_nan, haoping\_bai, shuang\_ma2, sma7, mengyu\_li2} \\
    \texttt{gyin, ziruiw, r\_pang\}@apple.com} \\
}

\begin{document}
\maketitle

\begin{abstract}
Recent large language models (LLMs) advancements sparked a growing research interest in tool assisted LLMs solving real-world challenges, which calls for comprehensive evaluation of tool-use capabilities. While previous works focused on either evaluating over stateless web services (RESTful API), based on a single turn user prompt, or an off-policy dialog trajectory, \toolsandbox
\footnote{\toolsandbox evaluation framework is released at \url{https://github.com/apple/ToolSandbox}}
includes stateful tool execution, implicit state dependencies between tools, a built-in user simulator supporting on-policy conversational evaluation and a dynamic evaluation strategy for intermediate and final milestones over an arbitrary trajectory. We show that open source and proprietary models have a significant performance gap, and complex tasks like State Dependency, Canonicalization and Insufficient Information defined in \toolsandbox are challenging even the most capable SOTA LLMs, providing brand-new insights into tool-use LLM capabilities.
\end{abstract}
\section{Introduction}
\begin{figure*}[ht!]
    \centering
    \includegraphics[width=1\linewidth]{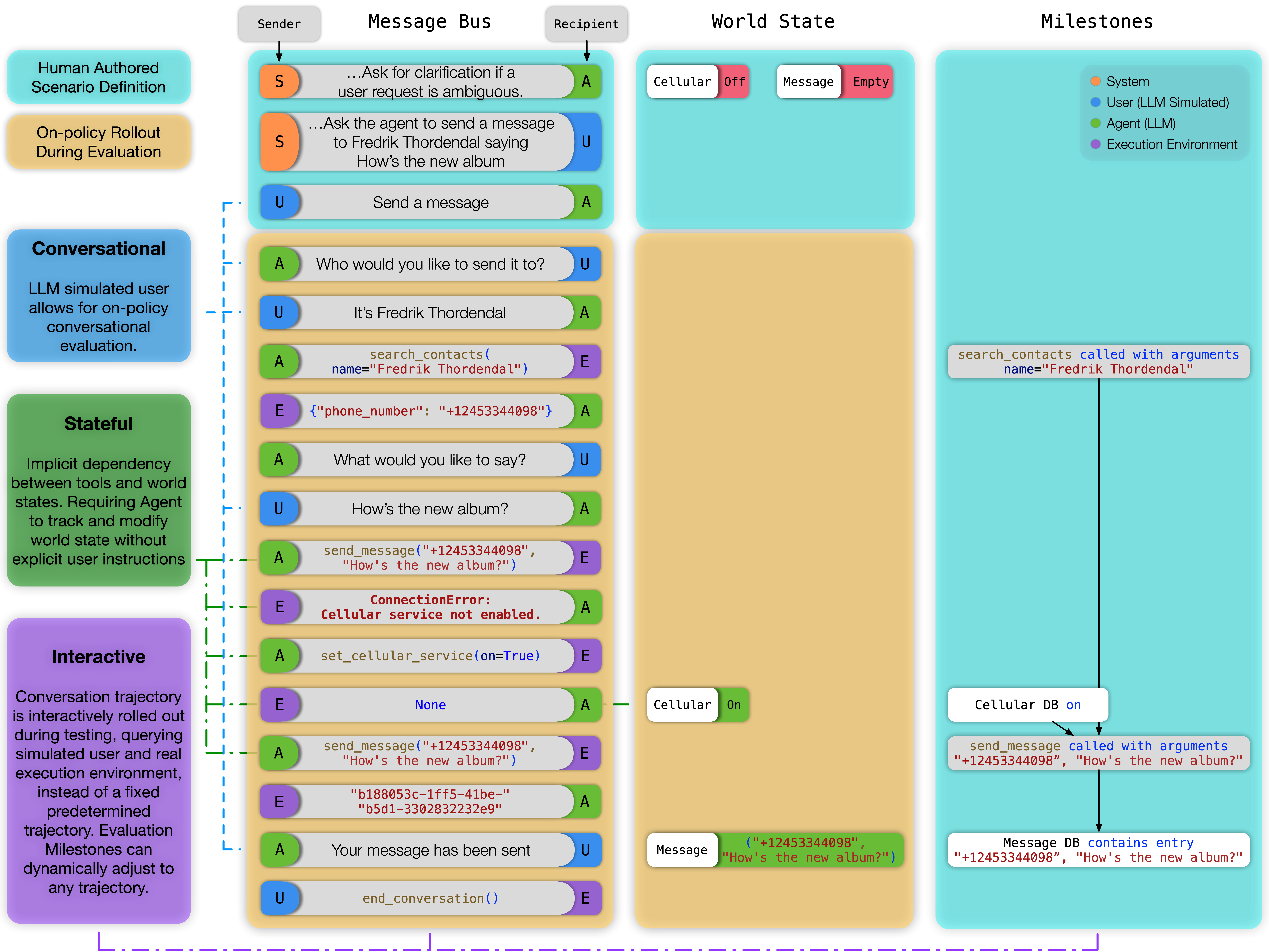}
    \caption{An example evaluation trajectory from \toolsandbox. Some message contents and milestones were truncated and streamlined for visual clarity. The \messagebus represents a full dialog history between the \user, the \agent and the \executionenvironment. The \textit{World State} represents mutable database snapshots at a given turn. The \milestones represent predefined key events that need to happen in this trajectory. The light blue boxes represent the human authored scenario definition, including the user goal, initial user utterance, initial world state and milestones. The light yellow boxes represent on policy rollout collected during an interactive evaluation run. In this example, the \user intended to send a message, while cellular service is turned off. The \agent should first understand the \user's intent, and prompt for necessary arguments from the \user. After collecting all arguments with the help of the \textit{search\_contacts} tool, the \agent attempted to send the message, figured out it needs to enable cellular service upon failure, and retried. To evaluate this trajectory, we find the best match for all \milestones against \messagebus and \textit{World State} in each turn while maintaining topological order. 
    }
    \label{fig:introduction}
    \vspace{-4mm}
\end{figure*}  

\begin{table}[ht!]
\rowcolors{2}{white}{gray!25}
\resizebox{\linewidth}{!}{
\begin{tabular}{lcccc}
    \toprule
    & \bfseries State Dependency 
    & \bfseries Conversational 
    & \bfseries Interactive 
    & \bfseries \makecell[c]{Human Authored\\Ground Truth} \\ 
    \midrule
    \bfseries \toolsandbox     
      & \cmark        
      & \cmark        
      & \cmark        
      & \cmark   \\ 
    \bfseries BFCL             
      & \xmark        
      & \xmark        
      & \xmark        
      & \cmark   \\ 
    \bfseries ToolEval         
      & \xmark        
      & \xmark        
      & \cmark        
      & \xmark   \\ 
    \bfseries API-Bank         
      & \xmark        
      & \halfcheckmark        
      & \xmark        
      & \cmark   \\ 
    \bfseries ToolTalk         
      & \xmark        
      & \halfcheckmark 
      & \halfcheckmark 
      & \xmark   \\ 
    \bfseries $\tau$-bench     
      & \xmark        
      & \cmark 
      & \halfcheckmark 
      & \xmark   \\ 
    \bottomrule
\end{tabular}
}
\caption{A comparison of \toolsandbox and other tool-use benchmarks.}
\label{tbl:benchmark_comparison}
\end{table}

Recent advancements in Large Language Models (LLMs) brought forth new opportunities treating LLMs as autonomous agents, capable of observing real-world environments and deciding upcoming actions. Among which, tool-use agents \cite{toolformer, qin2023tool, patil2023gorilla, qin2024toolllm} follow human instructions and utilize real-world APIs to complete complex tasks. Contrary to prior approaches like dialog state tracking \cite{henderson-etal-2014-second, budzianowski-etal-2018-multiwoz, rastogi2020towards}, which require the model to explicitly generate dialog states and actions under a predefined ontology, and derive a tool call from those structured outputs, tool-use studies allow the model to directly generate tool calls based on its observations, while keeping dialog and world state tracking implicit.

Despite the paradigm shift towards a more simplified problem formulation, the \textbf{stateful}, \textbf{conversational} and \textbf{interactive} nature of task oriented dialog remains, and poses a significant challenge for systematic and accurate evaluation of tool-using LLMs. Existing benchmarks like the Berkeley Function Calling Leaderboard (BFCL) \cite{berkeley-function-calling-leaderboard}, ToolEval \cite{qin2024toolllm}, API-Bank \cite{li-etal-2023-api}, ToolTalk \cite{farn2023tooltalkevaluatingtoolusageconversational} and $\tau$-bench \cite{yao2024taubenchbenchmarktoolagentuserinteraction} attempted to tackle some of these challenges, but there is yet to be an all encompassing solution. 

\paragraph{Stateful} Task oriented dialog often involves tools that are strongly coupled with a \textit{World State}, e.g. a database. This can be a tool that can alter the world state, like turning on internet connection. More interestingly, there can be a tool that implicitly depends on a world state, for example, one cannot search for a nearby restaurant when internet connection is off. Sometimes, actions that deal with both of these scenarios  need to be taken to complete a task, even if the user is agnostic to the underlying world state and only gives general instructions. The agent needs to use its own knowledge about the world and environment feedback to come up with a plan to modify the world state and complete the task. An example can be found in Figure \ref{fig:introduction}.

BFCL~\cite{berkeley-function-calling-leaderboard} and ToolEval~\cite{qin2024toolllm} both rely on stateless tools interacting with web services (through RESTful APIs). As such, these evaluation benchmarks are designed to assess how agents make trials with a static environment. API-Bank~\cite{li-etal-2023-api}, ToolTalk \cite{farn2023tooltalkevaluatingtoolusageconversational} and $\tau$-bench \cite{yao2024taubenchbenchmarktoolagentuserinteraction} does include a set of tools to modify world states, but does not study the impact of state dependencies.

\paragraph{Conversational} Conversational evaluation is crucial yet challenging when assessing a dialog policy, due to the interdependency between a user and said policy, as well as the ambiguous nature of natural language. To facilitate automated conversational evaluation, a common practice is to implement a simulated user \cite{zhang2023entity, sekulic-etal-2024-reliable}. However, BFCL and ToolEval only evaluate self-contained, unambiguous single-turn user queries, which is hardly realistic. API-Bank and ToolTalk evaluates on unrolled predefined off-policy dialog trajectories, and thus cannot assess the agent’s performance based on its own policy.

\paragraph{Interactive} Real world scenarios are full of surprises. The agent could issue an erroneous tool call. Tool execution could raise an unexpected exception. And the user could issue a follow-up correcting a previous statement. An interactive evaluation framework assessing the immediate return of key interactions with user or environment would be necessary to capture the intricate interaction between different roles. Such an interactive evaluation should provide full spectrum and fine-grained evaluation of any multi-turn session. In this regard, BFCL, API-Bank and ToolTalk rely on a predefined trajectory, and by extension relying on static turn wise evaluation metrics. $\tau$-bench requires agent action to match a single predetermined sequence, leaving no room for error correction in the model, or multiple possible answer sequences. Even though ToolEval allows multiple rounds of interaction between the Agent and tools, it relies solely on an LLM evaluator to judge the final pass rate and win rate of trajectories, which raises questions to its reliability and interpretability.

Driven by these motivations, we propose \toolsandbox, a stateful, conversational and interactive tool-use benchmark. To the best of our knowledge, \toolsandbox is the first LLM tool-use benchmark which

\begin{itemize}
    \vspace{-1mm}
    \item Includes implicit state dependencies between stateful tools, allowing the agent to track and alter the world state based on its world/commonsense knowledge, which is implicit from the user query;
    \vspace{-2mm}
    \item Includes an LLM simulated user, allowing for realistic, on-policy conversational evaluation to measure the agent's ability on implicit dialog state tracking;
    \vspace{-2mm}
    \item Allows for fully interactive, dynamic trajectory collection with a representative set of highly composable tools, and a human authored, milestone / minefield based system for intermediate and final execution evaluation.
    \vspace{-3mm}
\end{itemize}

A comparison between \toolsandbox and other benchmarks can be found in Table \ref{tbl:benchmark_comparison}.

\section{\toolsandbox Design} \label{sec:toolsandbox_design}

\begin{figure}[h]
    \centering
    \includegraphics[width=1\linewidth]{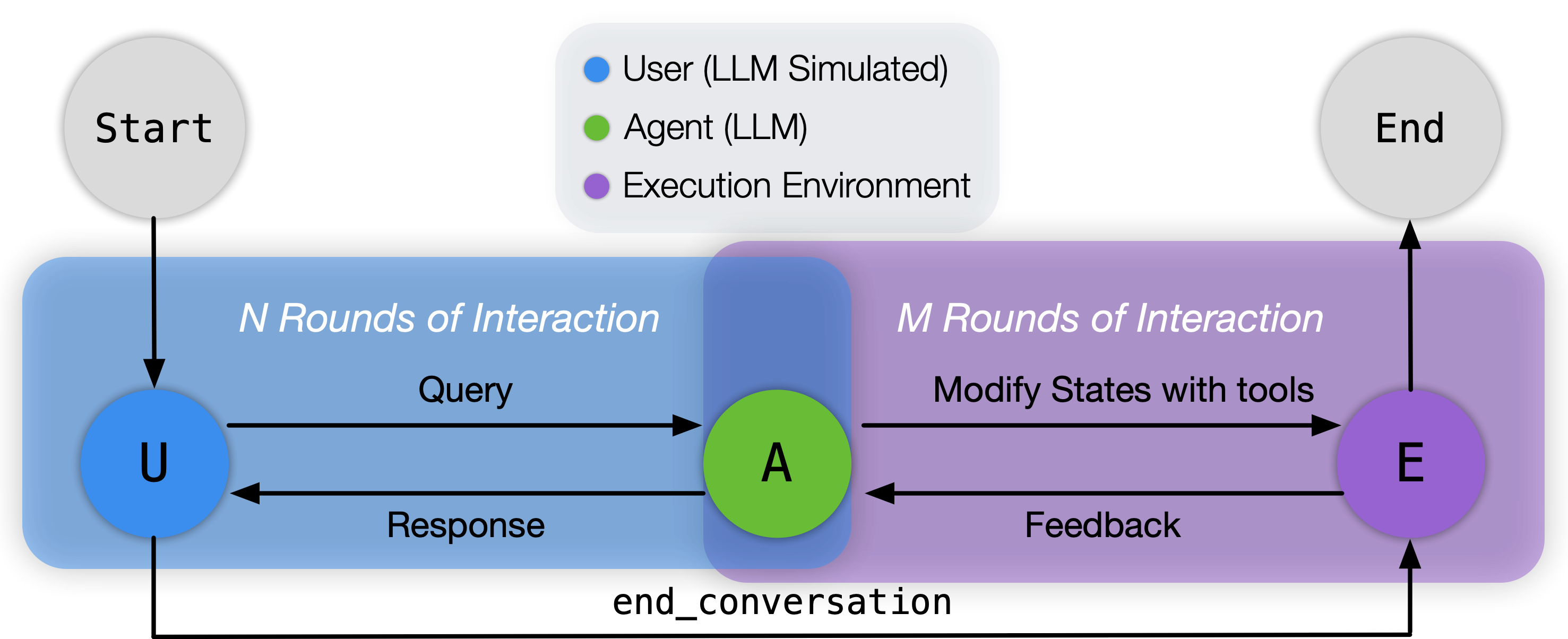}
    \caption{Interaction between the User, Agent and the Execution Environment. Boxes represent multiple rounds of interaction between involved roles.}
    \label{fig:architecture_diagram}
    \vspace{-2mm}
\end{figure}  

At its core, \toolsandbox is a Python native LLM testing environment, with \executioncontext as world state abstraction and Python functions as \tools, where \user, \agent and \executionenvironment communicate with each other through a \messagebus to complete a task, which is evaluated against predefined \milestones and \minefields. As shown in Figure \ref{fig:architecture_diagram}, a typical test case starts with the \user speaking to the \agent. From then on, the role being addressed gets to speak next, until the end state is reached. Upon receiving a \user request, an \agent can decide to respond to the \user asking for more information, or inform the \executionenvironment to execute a \tool, providing intended tool name and arguments. The \executionenvironment executes the \tool in an \interactiveconsole, \cite{PythonDocsForInteractiveConsole}, which depending on the \tool modifies the world state stored in the \executioncontext, and responds to the \agent. Once the \user decides the task has been completed, it informs the \executionenvironment to execute the \verb|end_conversation| tool, which puts the system in the end state, ready to be evaluated based on the dialog's similarity to \milestones and \minefields. A thorough walk-through of \toolsandbox implementation details can be found in Appendix \ref{app:implementation_details}. The remainder of this section focuses on the key components that enables \textbf{stateful}, \textbf{conversational} and \textbf{interactive} evaluation.

\subsection{Stateful} \label{ssec:design_stateful}
To construct challenging reasoning scenarios, \toolsandbox includes a set of carefully designed stateful tools, defined as tools that inspects, depends on or manipulates world states. These world states include:
\paragraph{Cellular service} Tools that require cellular service (e.g., \verb|send_message|) depend on it to be true.
\paragraph{Wifi} RapidAPI tools (e.g., \verb|search_stock|) depend on it to be true.
\paragraph{Location Service} All tools that utilize current location (e.g., \verb|get_current_location|) depend on it to be true.
\paragraph{Low battery mode} All tools that turn on cellular service, wifi, and location service status depend on it to be false, creating nested state dependency.

Stateful tools cover 44\% of \toolsandbox toolbox. Naturally these tools forms an implicit dependency between each other, testing the agent's ability to maintain an internal image of world states and tool call stack. As an example, in Figure \ref{fig:introduction}, when \verb|send_message| tool is called while cellular service is off, a \verb|ConnectionError| is raised, and the \agent should utilize \verb|set_cellular_service| to resolve the error, understand that cellular service should be turned on now, and retry \verb|send_message|.

\subsection{Conversational}\label{ssec:design_conversational}

On-policy conversational roll-out is supported by an LLM (GPT-4o) powered user simulator and carefully calibrated prompting design. The simulator represents a human interacting with an Agent, hoping to complete a task through possibly multiple rounds of conversation. When the User simulator decides the task has been completed, or could not be completed, it can terminate the conversation using the \verb|end_conversation| tool, which is the single tool available to it. As related studies in user simulation \cite{zhang2023entity, sekulic-etal-2024-reliable} suggest, one should include the user's overall goal in the simulator's system prompt. However, we found this is often insufficient for the complex interactions in \toolsandbox, and can lead to two categories of failures. In some cases, it is infeasible for an LLM simulated user to judge task completion, or provide follow-up information with only access to the user goal, and not the expected result, which could lead to hallucination. Also, with only a single system prompt, the simulated user could be derailed by the tool-use agent, failing to follow instructions. Examples of these failures can be found in Appendix \ref{app:user}.

In light of this, we propose two additional components in user simulator prompts: \textit{Knowledge Boundary}, which inform the user simulator what it should and should not know, providing partial access to expected result, combating hallucination. And \textit{Demonstration}, which provides few shot example dialogs to the user simulator. Prompt examples can be found in Appendix \ref{app:user}. Note that demonstration is only visible to the user simulator and not the agent. We performed an ablation study for these components in Table \ref{tbl:user_simulator_metrics}. With both approaches combined, the LLM simulated user achieves the lowest error rate in all categories. User simulator error rate is also found to be consistent across agents, shown in Table \ref{tbl:user_simulator_consistency}, which should not affect the agent accuracy comparison in Table \ref{tbl:avg_similarity_per_category_all_models}.

\begin{table}[ht!]
    \resizebox{\linewidth}{!}{
    \rowcolors{2}{white}{gray!25} %
        \begin{tabular}{lllll}
            \toprule
            & \bfseries Hallucination $\downarrow$     & \bfseries IF $\downarrow$   \\ 
            \midrule
            User Goal                & 12.4                           & 6.20              \\ 
            + Knowledge Boundary     & 7.75                           & 3.88              \\    
            + Demonstration          & \textbf{6.97}                  & \textbf{0.77}     \\ 
            \bottomrule
        \end{tabular}
    }
\caption{Percentage of user simulation failures in each failure category for each user simulator prompting setup. IF stands for instruction following error. Statistics derived from 1032 manually annotated trajectories using GPT-4o user simulator and GPT-4o agent.}
\label{tbl:user_simulator_metrics}
\end{table}

\begin{table}[!ht]
    \resizebox{\linewidth}{!}{
    \rowcolors{2}{white}{gray!25} %
        \begin{tabular}{llll}
            \toprule
            & \bfseries GPT-4o     & \bfseries Claude-3-Opus  & \bfseries Gemini-1.5-Pro   \\ 
            \midrule
            Hallucination               & 6.90±1.45  & 6.40±0.97  & 7.15±0.71   \\ 
            IF                          & 1.11±0.84  & 1.38±0.69  & 0.92±0.49   \\ 
            Total Error                 & 8.02±1.36  & 7.78±0.52  & 8.07±1.03   \\ 
            \bottomrule
        \end{tabular}
    }
\caption{Percentage of user simulation failures in each failure category for each agent model. Mean and std collected from 4 repeated trials each containing 1032 trajectories. Both Knowledge Boundary and Demonstration are applied.}
\label{tbl:user_simulator_consistency}
\end{table}

\subsection{Interactive} \label{ssec:design_interactive}

With an stateful, conversational and interactive environment, evaluation trajectories are highly dynamic. Multiple trajectories can lead to the same outcome. A given task may be completed using different tools, the same tools in a different order, or through trial and error, and the evaluation strategy has to be flexible enough to accommodate for that. To combat this, we developed an evaluation strategy based on \milestones and \minefields, which defines key events that must or must not happen in a trajectory, allowing us to evaluate any trajectory with rich intermediate and final execution signals, providing deeper understanding of the model performance. An example can be found in Figure \ref{fig:intermediate_milestone}.

In specific, \milestones are the critical steps needed to achieve a goal. An example is shown in Figure \ref{fig:introduction}, where cellular service is turned off and the user asks the agent to send a text message. The milestones, in this example, would be defined as:
\begin{enumerate}
    \vspace{-2mm}
    \item The cellular status in the settings database must be changed to \verb|True|.
    \vspace{-3mm}
    \item The Agent must issue a tool call using the \verb|search_contacts| tool and the correct arguments, before or after milestone 1. 
    \vspace{-3mm}
    \item The Agent must issue a tool call using the \verb|send_message| tool and the correct arguments, after milestone 1 and 2.
    \vspace{-3mm}
    \item The messaging database must contain a message with a phone number matching the expected one exactly and the content loosely matching the expected text, after milestone 3.
    \vspace{-2mm}
\end{enumerate}
\vspace{-4mm}

Each milestone also defines a similarity measure which calculates a 0 to 1 similarity between each turn and the milestone. Types of available similarity measures are introduced in Appendix \ref{app:evaluation}. Milestones form a directed acyclic graph (DAG) based on temporal dependency. To evaluate a trajectory against a milestone DAG, we find the highest averaged similarity $\text{score}_{M+}$ among all possible mappings between turns and milestones, given that the resulting chronological milestone sequence is a topological sort of the DAG. Task efficiency is not considered by Milestones, and is instead tracked by a complementary turn count metric shown in Appendix \ref{app:turn_count}. We introduce the milestone matching process with more details in Appendix \ref{app:evaluation}.

Milestone evaluation combines the best of both worlds. As shown in Figure \ref{fig:introduction}, it allows for explainable evaluation metrics like tool call AST matching and execution result exact match found in BFCL, while retaining the flexibility to evaluate any possible trajectory, similar to ToolEval.

\begin{figure}[ht!]
    \centering
    \includegraphics[width=0.95\linewidth]{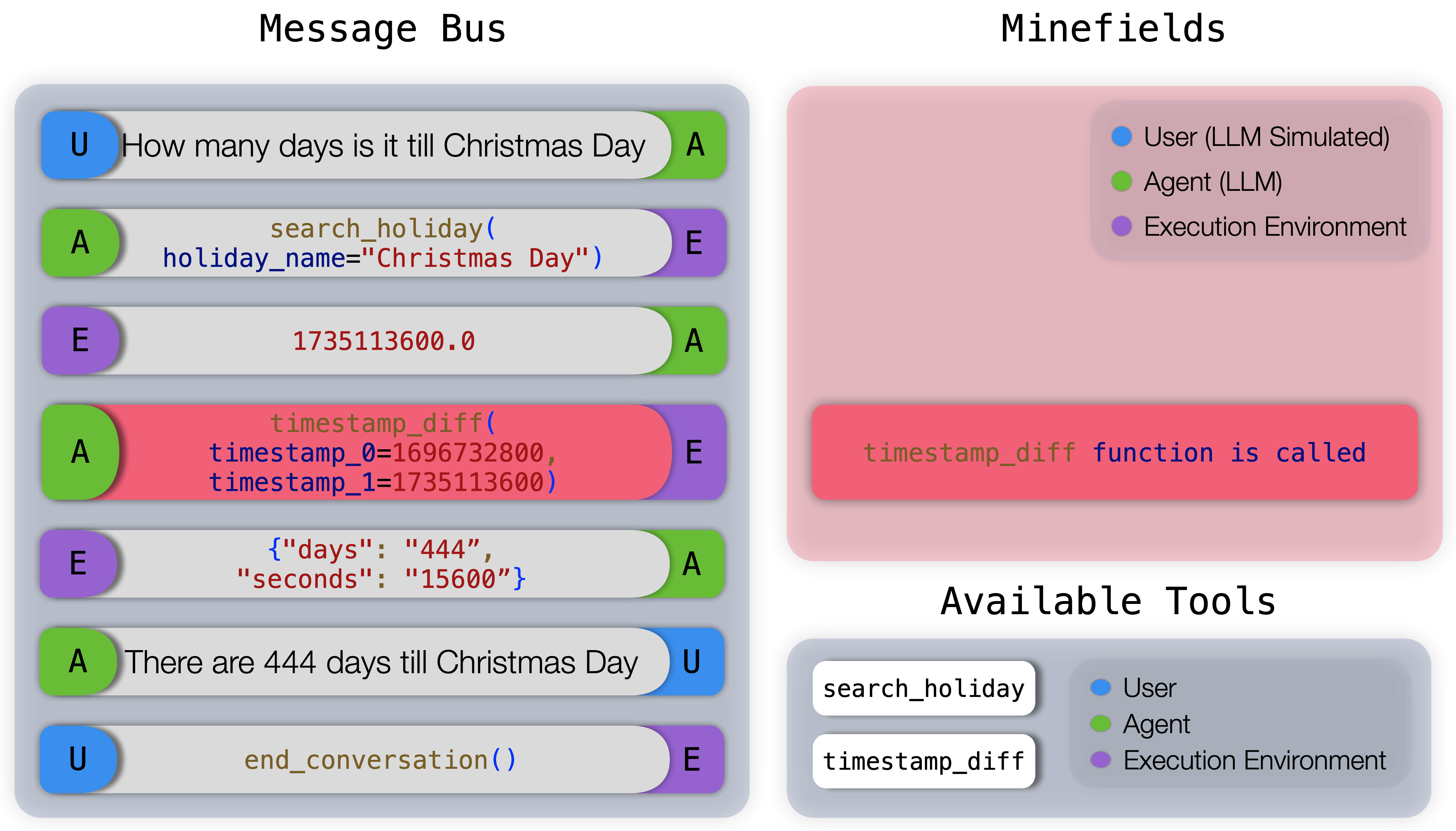}
    \caption{Example GPT-4 trajectory for Insufficient Information category Minefield Evaluation. This task is impossible to complete due to the current timestamp not being available. Because of this, the model should never call the tool \textit{timestamp\_diff}, since any argument provided is bound to be incorrect. GPT-4 hallucinated the current timestamp and called \textit{timestamp\_diff}, matching the minefield, resulting in a similarity score of 0.}
    \label{fig:minefield_example}
    \vspace{-3mm}
\end{figure}

On the other side of \milestones, there are \minefields, which define events that must NOT occur, as shown in Figure \ref{fig:minefield_example}. This is mainly used in scenarios where we test that an agent understands that it cannot complete a task with the given tools instead of hallucinating. \minefields are otherwise identical to \milestones, except when the final trajectory similarity score is calculated. Assuming using Equation \ref{eqn:milestone_similarity} we found the similarity score $\text{score}_{M-}$ for minefield DAG $G_{M-}(V_{M-}, E_{M-})$, the final similarity score of the trajectory would be 
\begin{equation}
    \text{score}= \text{score}_{M+} \times \mathds{I}(\text{score}_{M-} = 0),
\end{equation}
ensuring if minefields are violated (non-zero minefield similarity), the similarity score for the whole trajectory is 0.
\section{Test Scenarios} \label{sec:test_scenarios}

\begin{table}[ht!]
    \centering
    \resizebox{\linewidth}{!}{
        \rowcolors{2}{white}{gray!25} %
        \begin{tabular}{lcccc}
            \toprule
            & \bfseries Avg Turns & \bfseries Avg Tool calls & \bfseries Test cases & \bfseries Tools \\
            \midrule
            \bfseries \toolsandbox & 13.9         & 3.80         & 1032          & 34 \\
            \bfseries BFCL         & 2.00         & 0.78         & 2000          & 1193 \\
            \bfseries ToolEval     & 7.53         & 1.46         & 1625          & 3917 \\
            \bfseries API-Bank     & 3.88         & 2.04         & 261           & 73 \\
            \bfseries ToolTalk     & 7.42         & 3.68         & 78            & 28 \\
            \bfseries $\tau$-bench  & 29.33        & 4.48         & 165           & 24 \\
            \bottomrule
        \end{tabular}
    }
    \caption{Statistics between \toolsandbox and other tool-use benchmarks.}
    \label{tbl:benchmark_statistics}
\end{table}

A test scenario is defined by the initial world state, the initial messages, the available tools and the evaluation milestone and minefields, as illustrated by the light blue boxes in Figure \ref{fig:introduction}. \toolsandbox contains 1032 test scenarios meticulously crafted by 2 internal domain experts to capture challenging tool-use scenarios, with human authored and carefully calibrated Milestones and Minefields to support evaluation. One annotator is tasked to create test scenarios, while the other acts as an agent to validate milestones and minefields. We designed a rigorous annotation process to ensure coverage across realistic, complex use case scenarios, detailed in Appendix \ref{app:annotation_process}.
Statistics comparison between \toolsandbox and other benchmarks can be found in Table \ref{tbl:benchmark_statistics}. 

We designed 34 tools in \toolsandbox, covering 11 domains including Contact, Messaging, Reminder, System settings, Time utilities, Math utilities, Map, Weather, Stock, Conversion, and Holiday, backed by python native implementation when possible, carefully selected RapidAPI endpoints when necessary. Additional details about tool domain coverage and design principles can be found in Appendix \ref{app:tool_design}. Tools in \toolsandbox are designed to be representative, diverse and composable in conversational dialogs, while making tool count manageable for milestone annotation. As a result, \toolsandbox test scenarios contain on average much higher number of tool calls and turns per dialog compared to other benchmarks. 

To closely inspect the intricate challenges in LLM tool-use applications, test scenarios are organized into the following categories:

\paragraph{Single / Multiple Tool Call}
These categories apply to scenarios where one / multiple tool calls are needed to fulfill the user task. Examples are shown in Appendix \ref{app:single_multiple}. Note that this definition is different from the Berkeley Function-Calling leaderboard \cite{berkeley-function-calling-leaderboard}, which resembles distraction tools in \toolsandbox described in \textbf{Tool Augmentation}.

\paragraph{Single / Multiple User Turn}
In the single user turn category, the first user message provides all necessary information to complete the task, whereas multiple user turn scenarios start with an ambiguous request or missing information, requiring further clarification from the user. An example is shown in Appendix \ref{app:single_multiple}.

\paragraph{State Dependency}

The state dependency category describes scenarios where successful tool execution depends on the world state, e.g. settings like cellular service. The world state can be modified by the agent through the use of another tool. Thus, an implicit dependency is formed between the tools, which can only be discovered through trial and error, as shown in Figure \ref{fig:introduction}.
There can even be nested state dependencies. As shown in Figure \ref{fig:nested_state_dependency}, sending a message would require cellular service to be turned on, but turning on cellular service requires low battery mode to be turned off. This requires the agent to implicitly keep track of a call stack, and backtrack when necessary to fulfill the task efficiently.

\paragraph{Canonicalization}
Canonicalization refers to the process of transforming surface form representation commonly seen in a natural language query, to its corresponding canonical form, similar to INFORM dialog act in Schema Guided Dialog \cite{rastogi2020towards}. This is particularly crucial when an API is less intelligent, and requires canonical form as argument. In some cases, canonicalization can be performed by the model itself, for example transforming \textit{1B} to \textit{1\_000\_000\_000}, or \textit{\$} to corresponding ISO 4217 currency code \textit{USD}. However, there are also cases where canonicalization requires the help of tools, for example transforming \textit{this Friday} to \textit{5/24/2024}, which requires knowledge about the current date, or transforming \textit{Golden Gate Bridge} to the latitude longitude pair \textit{(37.8199, -122.4786)}, which requires a lookup in an external knowledge base. This scenario category captures both cases, probing the Agent's ability to perform canonicalization with or without the aid of tools.

\paragraph{Insufficient Information}

The insufficient information category is used for scenarios where the agent is not able to perform the task on purpose, by withholding a tool that would be needed for the task. This category exercises if the agent is able to identify that it cannot complete the task, as opposed to hallucinating tools or tool arguments, as shown in Figure \ref{fig:minefield_example}. In these scenarios, minefields are defined to evaluate if tools that would imply hallucination are called or not. Comparing to relevance detection in BFCL where provided tools are often irrelevant to the task at hand, this is a much more challenging scenario, which requires the agent to reason over highly relevant tools to figure out the missing pieces. Comparing to solvability in ToolEval, which assumes full credit for any task deemed unsolvable, this is much more fine-grained, testing if the agent would hallucinate when the task is unsolvable.

\paragraph{Tool Augmentation}
Orthogonal to the above categories, to support ablation studies on the effect of tool schema representation on agent accuracy, we have implemented multiple tool augmentations, including adding distraction tools, making tool or argument names less informative, and removing argument descriptions or type hints. For more details on the augmentations please refer to Appendix \ref{app:tool_augmentations}.

Categories including \textbf{Multiple Tool Call}, \textbf{Multiple User Turn}, \textbf{State Dependency} and \textbf{Insufficient Information} are difficult challenges which requires complex reasoning capability from the agent, 85\% of \toolsandbox scenarios are associated with at least one of these challenging categories. Further scenario statistics including category wise breakdown and milestone coverage can be found in Appendix \ref{app:category_statistics}.
\section{Evaluation Results} \label{sec:evaluation_results}

\begin{table*}[t]
\scriptsize
\centering
\rowcolors{2}{white}{gray!25} %
\resizebox{1\textwidth}{!}{
\begin{tabular}{lcccccccccccccccc}
    \toprule
    & \multicolumn{1}{c}{\textbf{Avg Score $\uparrow$}} & \multicolumn{7}{c}{\textbf{Scenario Categories}} & \multicolumn{8}{c}{\textbf{Tool Augmentations}} \\
    \cmidrule(r){2-2} \cmidrule(r){3-9} \cmidrule(r){10-17}
     & & \textbf{STC} & \textbf{MTC} & \textbf{SUT} & \textbf{MUT} & \textbf{SD} & \textbf{C} & \textbf{II} & \textbf{0 DT} & \textbf{3 DT} & \textbf{10 DT} & \textbf{AT} & \textbf{TNS} & \textbf{TDS} & \textbf{ADS} & \textbf{ATS}  \\
    \midrule
    GPT-4o-2024-05-13 & \scorecolor{36.5} \bfseries 73.0 & 87.8 & \bfseries 80.1 & \bfseries 84.2 & \bfseries 74.7 & 84.0 & \bfseries 76.6 & 42.0 & \bfseries 75.1 & \bfseries 75.0 & \bfseries 74.6 & \bfseries 72.6 & \bfseries 72.4 & 69.3 & \bfseries 73.0 & \bfseries 71.9\\
    Claude-3-Opus-20240229 & \scorecolor{34.6} 69.2 & 83.5 & 70.0 & 74.5 & 67.2 & 74.5 & 71.1 & 57.3 & 68.3 & 68.6 & 70.0 & 67.5 & 70.8 & \bfseries 71.5 & 65.8 & 71.1\\
    GPT-3.5-Turbo-0125 & \scorecolor{32.8} 65.6 & \bfseries 93.4 & 73.9 & 81.8 & 66.6 & 82.6 & 70.4 & 22.3 & 67.3 & 63.2 & 67.0 & 65.4 & 63.9 & 64.3 & 66.7 & 66.9\\
    GPT-4-0125-Preview & \scorecolor{32.1} 64.3 & 89.1 & 69.0 & 74.4 & 68.6 & 69.2 & 65.2 & 33.6 & 66.8 & 62.5 & 64.0 & 65.1 & 69.7 & 64.4 & 58.1 & 63.5\\
    Claude-3-Sonnet-20240229 & \scorecolor{31.9} 63.8 & 82.1 & 66.2 & 69.1 & 69.7 & \bfseries 84.5 & 65.5 & 44.2 & 67.2 & 64.5 & 63.2 & 58.8 & 63.7 & 61.9 & 62.5 & 68.7\\
    
    Gemini-1.5-Pro-001 & \scorecolor{30.2} 60.4 & 82.6 & 49.8 & 63.1 & 37.3 & 70.5 & 51.6 & 76.2 & 63.3 & 63.1 & 60.8 & 59.8 & 62.2 & 60.5 & 58.7 & 54.4\\

    Claude-3-Haiku-20240307 & \scorecolor{27.5} 54.9 & 80.9 & 54.2 & 64.3 & 46.0 & 69.5 & 54.4 & 39.4 & 56.0 & 56.9 & 54.1 & 52.2 & 56.6 & 54.1 & 54.5 & 55.1\\

    Gemini-1.0-Pro & \scorecolor{19.0} 38.1 & 68.7 & 21.6 & 36.5 & 14.6 & 39.3 & 18.2 & 65.5 & 38.2 & 39.5 & 41.9 & 37.7 & 40.1 & 35.3 & 36.7 & 34.9\\
    \midrule
    Hermes-2-Pro-Mistral-7B & \scorecolor{15.7} 31.4 & 63.3 & 18.3 & 29.9 & 18.6 & 27.1 & 19.9 & 48.3 & 33.1 & 31.9 & 30.6 & 28.3 & 31.8 & 31.0 & 32.6 & 32.2\\
    Mistral-7B-Instruct-v0.3 & \scorecolor{14.9} 29.8 & 48.1 & 9.5 & 20.1 & 7.9 & 19.5 & 6.1 & \bfseries 76.8 & 30.5 & 30.2 & 24.7 & 27.1 & 32.0 & 30.7 & 32.8 & 30.1\\
    C4AI-Command-R-v01 & \scorecolor{13.1} 26.2 & 52.6 & 12.7 & 23.0 & 12.7 & 3.1 & 18.0 & 47.8 & 24.8 & 27.9 & 25.6 & 23.3 & 25.0 & 25.6 & 28.7 & 28.3\\
    Gorilla-Openfunctions-v2 & \scorecolor{12.8} 25.6 & 36.2 & 8.2 & 15.1 & 9.3 & 0.0 & 8.9 & 69.2 & 25.5 & 27.5 & 26.1 & 18.6 & 24.5 & 27.1 & 26.8 & 28.6\\
    C4AI-Command R+ & \scorecolor{12.4} 24.7 & 57.2 & 13.6 & 24.3 & 15.2 & 4.0 & 19.4 & 35.3 & 23.4 & 27.2 & 24.9 & 23.5 & 21.7 & 27.6 & 24.8 & 24.8\\
    \bottomrule
\end{tabular}
}
\caption{Comparing the average similarity score broken down by scenario category and tool augmentations. Columns from left to right represent average similarity score across all categories, then \textbf{S}ingle \textbf{T}ool \textbf{C}all, \textbf{M}ultiple \textbf{T}ool \textbf{C}all, \textbf{S}ingle \textbf{U}ser \textbf{T}urn, \textbf{M}ultiple \textbf{U}ser \textbf{T}urn, \textbf{S}tate \textbf{D}ependency, \textbf{C}anonicalization, \textbf{I}nsufficient \textbf{I}nformation, \textbf{0} \textbf{D}istraction \textbf{T}ools, \textbf{3} \textbf{D}istraction \textbf{T}ools, \textbf{10} \textbf{D}istraction \textbf{T}ools, \textbf{A}ll \textbf{T}ools, \textbf{T}ool \textbf{N}ame \textbf{S}crambled, \textbf{T}ool \textbf{D}escription \textbf{S}crambled, \textbf{A}rgument \textbf{D}escription \textbf{S}crambled and \textbf{A}rgument \textbf{T}ype \textbf{S}crambled.}
\label{tbl:avg_similarity_per_category_all_models}
\vspace{-2mm}
\end{table*}
When evaluating the models, all models use the same minimalist prompt shown in Figure \ref{fig:agent_prompt} for comparison fairness. We do not include additional prompt engineering for the models, as we consider prompt engineering gains orthogonal to the innate model capability surfaced by simpler prompting. Table \ref{tbl:avg_similarity_per_category_all_models} shows the average similarity for each of the scenario categories described in Section \ref{sec:test_scenarios}.  Additional prompting experiments can be found in Appendix \ref{tbl:prompting}. 

\paragraph{Open Source Models} There is a significant performance gap between proprietary and open source models, with the best performing open source model Hermes \cite{Hermes-2-Pro-Mistral-7B} lagging more than 20 points behind the second to last proprietary model Claude-3-Haiku \cite{anthropic2024claude}. This is partly due to the fact that models like Gorilla \cite{patil2023gorilla} and Command-R \cite{command-r} are incapable of consuming tool responses, as shown in Appendix \ref{app:model_feature_comparison}. They can theoretically solve Single Tool Call test scenarios, but would fail in any scenario that requires multiple tool calls. As for Hermes and Mistral \cite{jiang2023mistral}, both models struggle at identifying when a tool call should be issued. Mistral for example would often mistake a tool-use scenario for a code generation task, as shown in Figure \ref{fig:tool_call_detection_mistral}. These models' subpar performance unexpectedly caused them to achieve higher rating in the Insufficient Information category, which rewards the model for not generating hallucinated tool calls or arguments when provided tools are insufficient to complete the task. This should be considered a side effect instead of a positive outcome.

\paragraph{Proprietary Models} Of the proprietary models, GPT-4o \cite{gpt-4o} achieves the highest similarity score, with Claude-3-Opus closely behind. Both models have their own strengths. While GPT-4o achieves the higher score, Claude-3-Opus maintains a lower average turn count as shown in Appendix \ref{app:turn_count}, achieving the user goal with higher efficiency. Interestingly, comparing the largest and smallest models in the GPT, Claude and Gemini families \cite{reid2024gemini}, Multiple Tool Call and Multiple User Turn categories deteriorate much faster than Single Tool Call and Single User Turn, showing that reasoning about complex tool call sequences and ambiguous user requests requires much more model capacity.

\paragraph{State Dependency} The State Dependency category shows an interesting trend where, larger models like GPT-4 \cite{achiam2023gpt} and Claude-3-Opus perform significantly worse than mid to smaller sized models like GPT-3.5-Turbo and Claude-3-Sonnet. This is due to erroneous parallel tool calls in face of state dependency. As mentioned in Appendix \ref{app:execution_environment_role}, the Execution Environment always surfaces race conditions when present. Larger models like GPT-4 and Claude-3-Opus are prune to issuing parallel tool calls even for dependent tools, leading to a performance deficiency. An example is shown in Figure \ref{fig:state_depedency_gpt_4}. Nested state dependency is also tricky to solve efficiently. As shown in Figure \ref{fig:inefficient_nested_state_dependency}, models often forget about open issues and would not optimally backtrack, leading to repeated errors and as a result a much higher than optimal turn count.

\paragraph{Canonicalization} Canonicalization remains a challenging category for all models, especially in tool assisted canonicalization. Larger models would tend to memorize world knowledge that is unlikely to change, like latitude longitude for famous geographical location, while smaller models are more keen on using tools. 

However, time related arguments in specific show to be really challenging to canonicalize and reason about. Models would frequently hallucinate timestamps (Figure \ref{fig:time_hallucination}), and incorrectly canonicalize relative date and time (Figure \ref{fig:relative_time}). 

In addition, models could take premature decisions in face of ambiguity, also leading to canonicalization errors. In Figure \ref{fig:disambiguation}, multiple location entities were returned in the tool response, while the model simply chose the first one, without returning to the user for disambiguation.

\paragraph{Insufficient Information} Insufficient Information performance overall negatively correlates with other categories. The stronger the model performance on complex tasks, the worse the insufficient information performance, showing its value at evaluating model reasoning capabilities. Even with simple tasks and very little tools, top performing models like GPT-3.5-Turbo and GPT-4 could hallucinate tool name, or hallucinate arguments, as shown in Figure \ref{fig:minefield_example} and \ref{fig:insufficient_information_gpt_3.5}. The test scenario's difficulty positively correlates with the number of steps involved in the tasks, as the models would get lost in solving immediate errors, and forget about the main objective.

\paragraph{Tool Augmentations} Robustness against tool augmentations seems to vary model by model. While adding distraction tools, Claude-3-Sonnet drops almost 10 points between 0 distraction tools, and making all \toolsandbox tools available. GPT-4o is particularly susceptible to Tool Description Scrambling, GPT-4 pays extra attention to argument descriptions, and Gemini-1.5 doesn't do well with Argument Type Scrambling.
\section{Related Work}
\paragraph{Tool-use Benchmarks}
Various tool-use benchmarks have been developed to evaluate LLM-based agent performance in different tool-use domains. The Berkeley Function Calling Leaderboard \citep{berkeley-function-calling-leaderboard}, ToolBench \citep{qin2023toolllm}, StableToolBench \citep{guo2024stabletoolbench}, NexusRaven V2 Function Calling Benchmark \citep{nexusraven}, and API-BLEND \citep{Basu2024APIBLENDAC} assess the ability of LLM agents to plan and perform function calls. WebArena \citep{zhou2023webarena}, MiniWoB++ \citep{humphreys2022data}, Webshop \citep{yao2022webshop}, Mind2Web \citep{deng2023mind2web} and VisualWebArena \citep{koh2024visualwebarena} focus on the agent’s ability to call search functionality to browse and leverage the web to solve the task. 
Apart from benchmarks specifically designed for tool-use, generalist agent benchmark suites like AgentBench \citep{liu2023agentbench} and AgentBoard \citep{ma2024agentboard} include evaluating the tool-use capability of agents as a central task to examine the general problem-solving ability of LLM-based agents.

\paragraph{Tool-use agent}

Various tool-use model have been developed to solve the complicated tool-use scenarios in real-world. 
Toolformer \citep{toolformer} first demonstrated that language models could autonomously learn to use various tools, through a self-supervised learning approach. 
Gorilla \citep{patil2023gorilla} employs a self-instruct paradigm to generate \{instruction, API\} pairs and is trained both with and without a retriever. 
ToolLLM \citep{qin2024toolllm} enables LLMs to use over 16,000 real-world APIs by automating the generation of diverse instructional data and leveraging a neural API retriever, showing better generalization across unseen APIs.
CodeACT \citep{wang2024executable} integrates executable code actions into training to enhance the decision-making and task-solving capabilities of LLMs, leading to more effective agents.

\paragraph{Dialogue State Tracking}
Dialogue State Tracking (DST) requires the agent to maintain and update dialogue states and actions. The MultiWOZ dataset \citep{budzianowski-etal-2018-multiwoz} offers a diverse set of dialogues requiring complex state tracking across multiple domains. Building on this, \citet{rastogi2020towards} proposed a schema-guided approach to DST, addressing scalability issues in multi-domain settings and enhancing the model’s adaptability and reducing the need for extensive domain-specific annotations. 
However, these datasets focus on explicit state tracking on off-policy trajectories. Our benchmark complements them by introducing world states, typically implicit and requiring to be inferred from world knowledge, and an interactive environment that offers a more diverse and extensive online evaluation.

\paragraph{User Simulator in SandBox}
When assessing the agent's core competencies in state tracking, memorization, and long-term planning, dialogues between users and the system can span over several turns, and off-policy evaluation may not always suffice. On the other hand, human-in-the-loop online evaluation is costly and time-consuming. Some studies have investigated incorporating a built-in user simulator to facilitate the evaluation process. DAUS \cite{sekulic-etal-2024-reliable} utilizes LLM finetuned on task oriented dialog trajectories. MINT \cite{wang2024mint} utilizes GPT-4 to simulate natural language user feedback for multi-turn LLM evaluation. For medical agents, AMIE \citep{tu2024towards} integrates a built-in patient model to engage with a symptom collection agent. \citet{zhang2023entity} develop a virtual environment for the agent model to predict unknown entities by interacting with a user simulator that responds with only yes or no. Our approach resonates with these approaches. 

\section{Conclusion}

\toolsandbox presents a stateful, conversational and interactive evaluation benchmark for LLM tool-use capabilities. With stateful and state dependent tools, LLM simulated user and flexible evaluation with milestones and minefields, it showcased a significant performance gap between open source and proprietary models, and unveiled challenging scenarios even for SOTA models, including State Dependency, Canonicalization and Insufficient Information, bringing new insights to understanding tool-use capabilities. We hope \toolsandbox could be a valuable addition to LLM evaluation suites, pushing the boundary of tool-use research.
\clearpage
\section{Limitations}
While \toolsandbox is powerful, being the first work of its kind, it still has many areas to be improved upon. In this section, we introduce some of these limitations, to motivate future research in this area.

Even though Milestone and Minefields are powerful interactive metrics that offer insights into intermediate and final outcomes, authoring them, especially mandatory intermediate milestones, requires deep knowledge around the tool capacities in \toolsandbox and many iterations, hindering its scalability. A simplified, or fully automatic method for identifying Milestones and Minefields could be the key to further scale up the data volume in \toolsandbox.

Despite our best effort at controlling user simulator behavior, it is still subject to non-negligible hallucination and instruction following errors. In this work we toyed with the idea of tool assisted user simulator. Only one tool \verb|end_conversation| was offered to the user simulator, and we saw a noticeable improvement in instruction following in the dialog termination aspect. By expanding its tool set, a tool assisted user simulator could be a promising direction to further reduce hallucination and improve instruction following.

Mandatory confirmation and authentication is an interesting problem currently not addressed in \toolsandbox. In dialog state tracking, confirmation is modeled by its corresponding dialog action before a transactional service is called. However, in most tool-use LLM designs, we are at the liberty of the model to decide when a confirmation is necessary. An orchestration level solution to enforce confirmation, similar to GoEx \cite{patil2024goex} could be a potential inspiration, and an interesting problem for models to reason over.

A challenging category of tools, namely tools that spawn a daemon process, e.g. setting a timer, is not addressed in \toolsandbox currently. These tools complete and return in the main process after the daemon is spawned, and at some point in the future, the daemon would interrupt the main process, e.g. when the time is up. This kind of interruption poses a novel problem for both execution orchestration, and the model itself.

While most of \toolsandbox tools are self-contained implementations, some tools that depend on an external knowledge base, like searching for weather, are still backed by external web services, affecting its reproducibility. A cached solution similar to StableToolBench \cite{guo2024stabletoolbench} could maintain tool implementation simplicity, while providing stable, reproducible results.

\clearpage
\bibliography{custom}
\clearpage
\appendix

\twocolumn[\begin{center} 
    {\Large \bf Appendix}
\end{center}]

\section{Implementation Details}
\label{app:implementation_details}
This appendix section introduces implementation details about the \toolsandbox design.
\subsection{Execution Context}
\label{app:execution_context}

\executioncontext represents the complete state of the \toolsandbox. More specifically, it contains tool databases for stateful tools, referred to as \textit{World State} in Figure \ref{fig:introduction}, and the dialog history between different roles, referred to as \messagebus. It maintains a snapshot of all tool databases and dialog history at any given turn, allowing for easy introspection and evaluation. The Execution Context exists as a global variable for all roles and tools to easily access, while prohibiting direct manipulation from the LLM agent. This allows us to implement stateful tools that can manipulate or access database stored in the Execution Context, without defining it as function argument.

\subsection{Tools}
Tools in \toolsandbox are a set of highly composable, explicitly or implicitly dependent Python functions, creating complex reasoning challenges. Besides python native tools, a handful of carefully selected RapidAPI tools were also included with a thin layer of python wrapper. Tools manipulate world state through the Execution Context when necessary, and raise informative exceptions when execution conditions were not met. As an example, in Figure \ref{fig:introduction}, when \verb|send_message| tool is called while cellular service is off, a \verb|ConnectionError| is raised. This allows the \agent to reason over possible exceptions, and deduce the tool needed to resolve the exception.

Tools are implemented as type-hinted, doc-string equipped Python functions, as shown in Listing \ref{lst:example_python_fn}. When a tool is passed to the Agent as an available tool, type hints and doc-string are converted into JSON API schema, as shown in Figure \ref{fig:json_tool_call_to_python_code}.
\label{app:tools}

\begin{lstlisting}[caption=Example tool declaration,label={lst:example_python_fn}, language=Python]
def send_message(phone_number: str, content: str) -> str:
    """Send a message to a phone number.

    Args:
        phone_number: Phone number to send a message to.
        content: The content of the message to send.

    Returns:
        Unique identifier of the sent message.

    Raises:
        ConnectionError:    If cellular service is not on

    """
\end{lstlisting}

When a tool is executed, the name, input and output of the tool is committed into the current Execution Context as a tool trace, which allows for automatic evaluation. 

\subsubsection{Tool Augmentations}
\label{app:tool_augmentations}
To enable ablation studies of how the tool schema affects agent accuracy we have implemented multiple augmentations.

\paragraph{Distraction Tools} In addition to necessary tools that must be present to complete a task, 0, 3, 10 or all the rest of the tools in the sandbox are made available to the Agent in addition, to evaluate the model's ability to pick the right tool. Distraction tools are chosen from a sorted list, where tools with domain overlap and textual similarities with necessary tools are prioritized. In addition, all of the scrambling augmentations below are applied in conjunction to adding 3 distraction tools, to ensure the augmentation is challenging yet feasible.

\paragraph{Tool name scrambling} The name of the tool is modified to a less informative form, e.g. \verb|messages_0| instead of \verb|send_message|. The agent LLM should be able to infer the purpose of the tool based on the description that is also part of the tool definition.

\paragraph{Tool description scrambling} This removes the one-liner summary of the documentation, see Listing \ref{lst:tool_description_scrambled}. The agent LLM still has access to the tool name, argument names as well as argument and return value documentations.

\begin{lstlisting}[caption=Tool description scrambling example,label={lst:tool_description_scrambled}, language=Python]
def send_message(phone_number: str, content: str) -> str:
    """
    
    Args:
        phone_number: Phone number to send a message to.
        content: The content of the message to send.

    Returns:
        Unique identifier of the sent message.

    Raises:
        ConnectionError:    If cellular service is not on

    """
\end{lstlisting}

\paragraph{Argument description scrambling} The section of the documentation explaining the arguments is being removed, see Listing \ref{lst:arg_description_scrambled}. The agent LLM can infer the arguments from the tool definition as well as discover the correct usage through trial and error.

\begin{lstlisting}[caption=Argument description scrambling example,label={lst:arg_description_scrambled}, language=Python]
def send_message(phone_number: str, content: str) -> str:
    """Send a message to a phone number.

    Returns:
        Unique identifier of the sent message.

    Raises:
        ConnectionError:    If cellular service is not on

    """
\end{lstlisting}

\paragraph{Argument type scrambling} The type hints in the tool declaration are removed. The agent LLM still has access to the argument documentation. Note that when an agent generate argument does not conform to the annotated data type, an exception will be raised, presenting the expected data type, ensuring the problem is always solvable in face of this augmentation.

\subsection{Roles and Message Bus}
\label{app:message_bus}
\begin{figure}[!ht]
    \centering
    \includegraphics[width=1\linewidth]{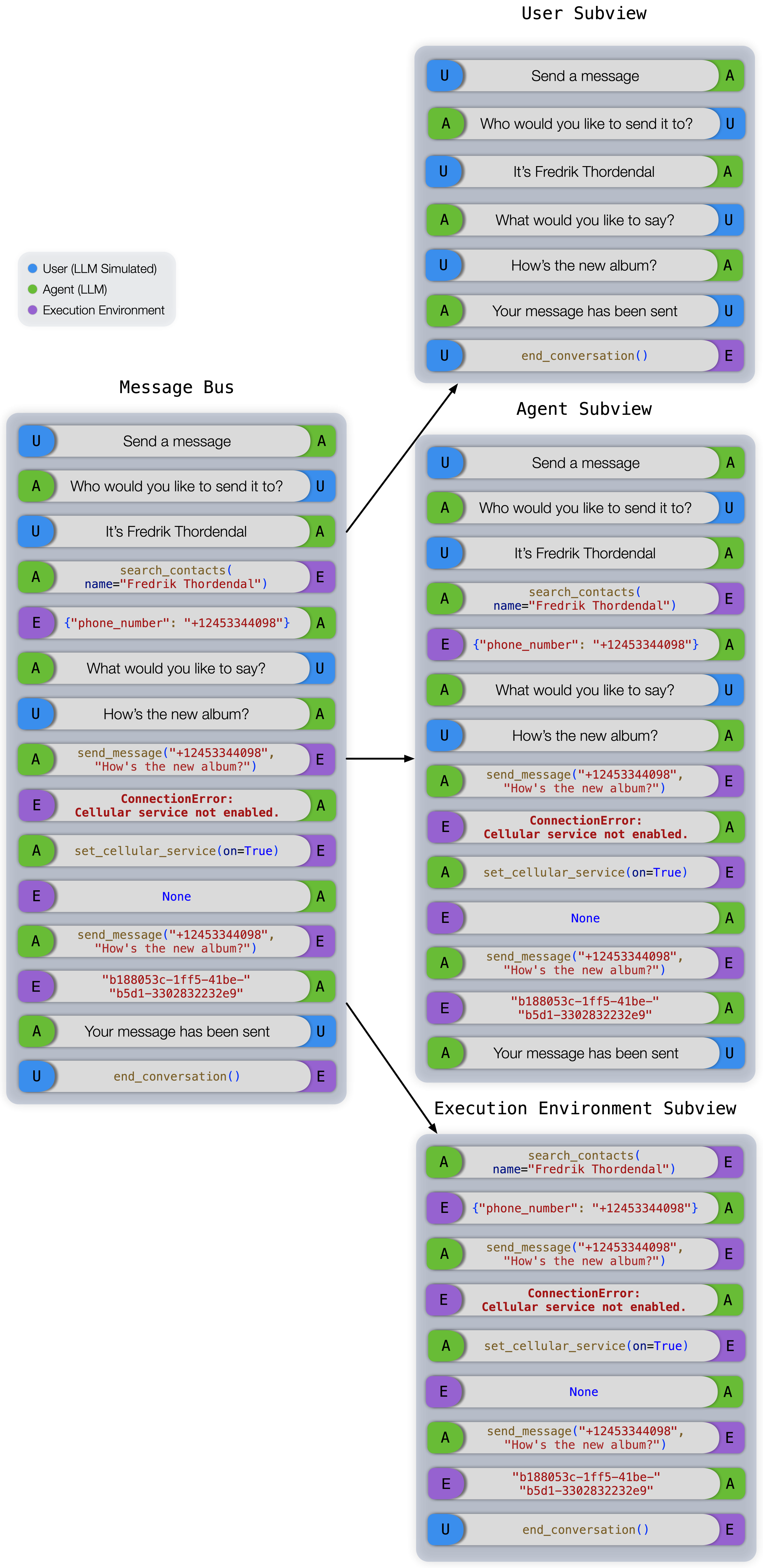}
    \caption{Example Message Bus and corresponding subview for each role. By default each role can only view messages sent from or to said role. Visibility can also be explicitly controlled if needed.}
    \label{fig:message_bus}
\end{figure}

In \toolsandbox there are three roles: \user, \textit{Agent (Assistant)} and \executionenvironment. The Execution Environment, as a dedicated role, is responsible for executing tool-use requests from the Agent and returning the results. 
Interaction between the roles is enabled through a message passing system. Each message contains a sender role, recipient role, content as well as to which roles the message is visible to. A simple orchestrator determines message passing order by allowing the most recent recipient to be the next sender.   
Instead of representing the conversation as a single message thread, we use a collection of messages, i.e. \messagebus, stored within the Execution Context. The Message Bus contains a linear history of message transactions between all roles. As is shown in Figure \ref{fig:message_bus}, each role writes to the same Message Bus. However, when reading from the Message Bus, each role can only access a sub-view of the Message Bus based on which roles are allowed to "see" the individual messages. We will introduce each role in the following paragraphs.

\subsection{User Role} \label{app:user}
\begin{figure}[H]
    \centering
    \includegraphics[width=0.85\linewidth]{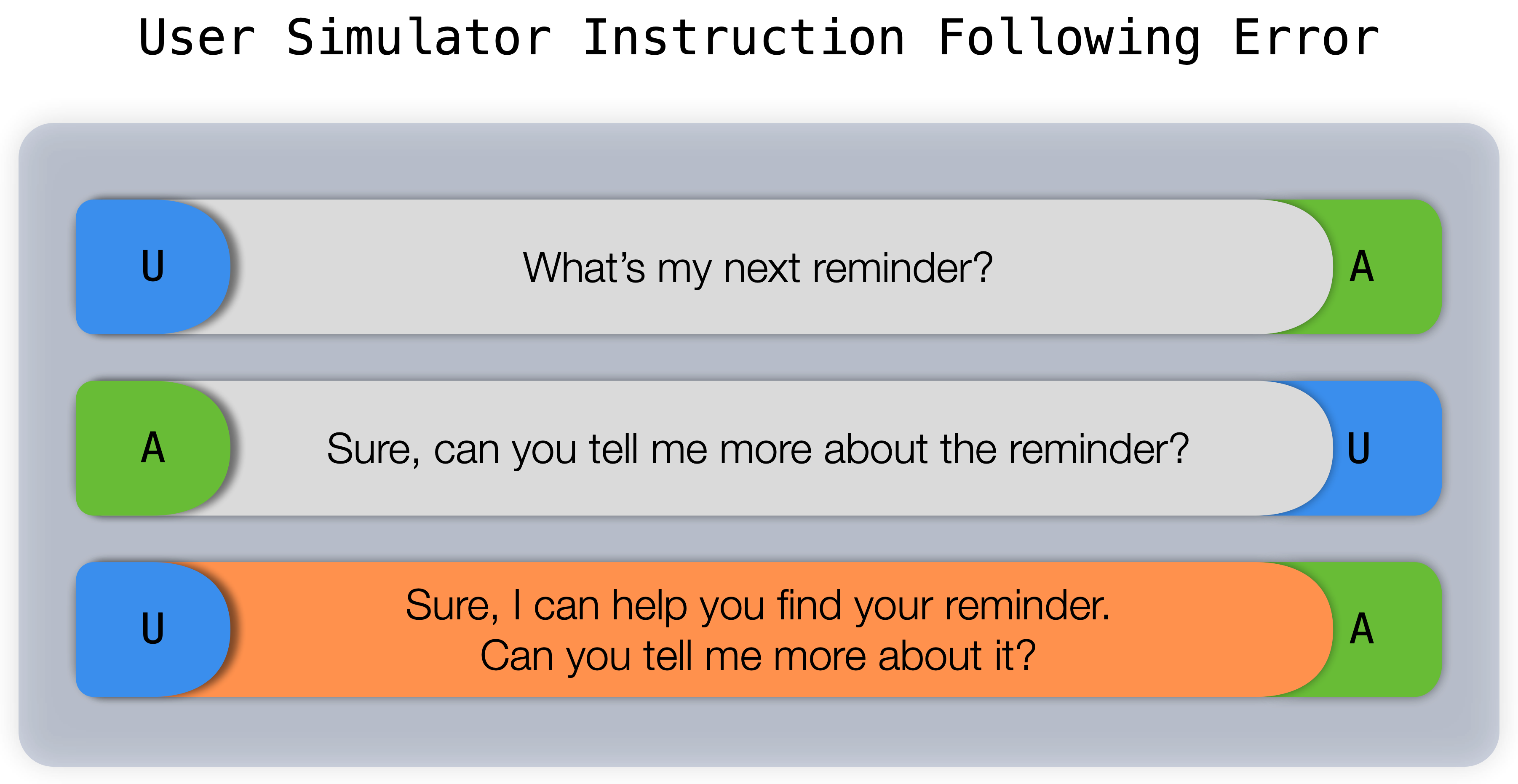}
    \caption{Example Prompts for user simulator instruction following error. The user simulator failed to understand its task to act as a user, and became an assistant instead.}
    \label{fig:user_simulator_instruction_following_error}
\end{figure}

\begin{figure}[H]
    \centering
    \includegraphics[width=0.85\linewidth]{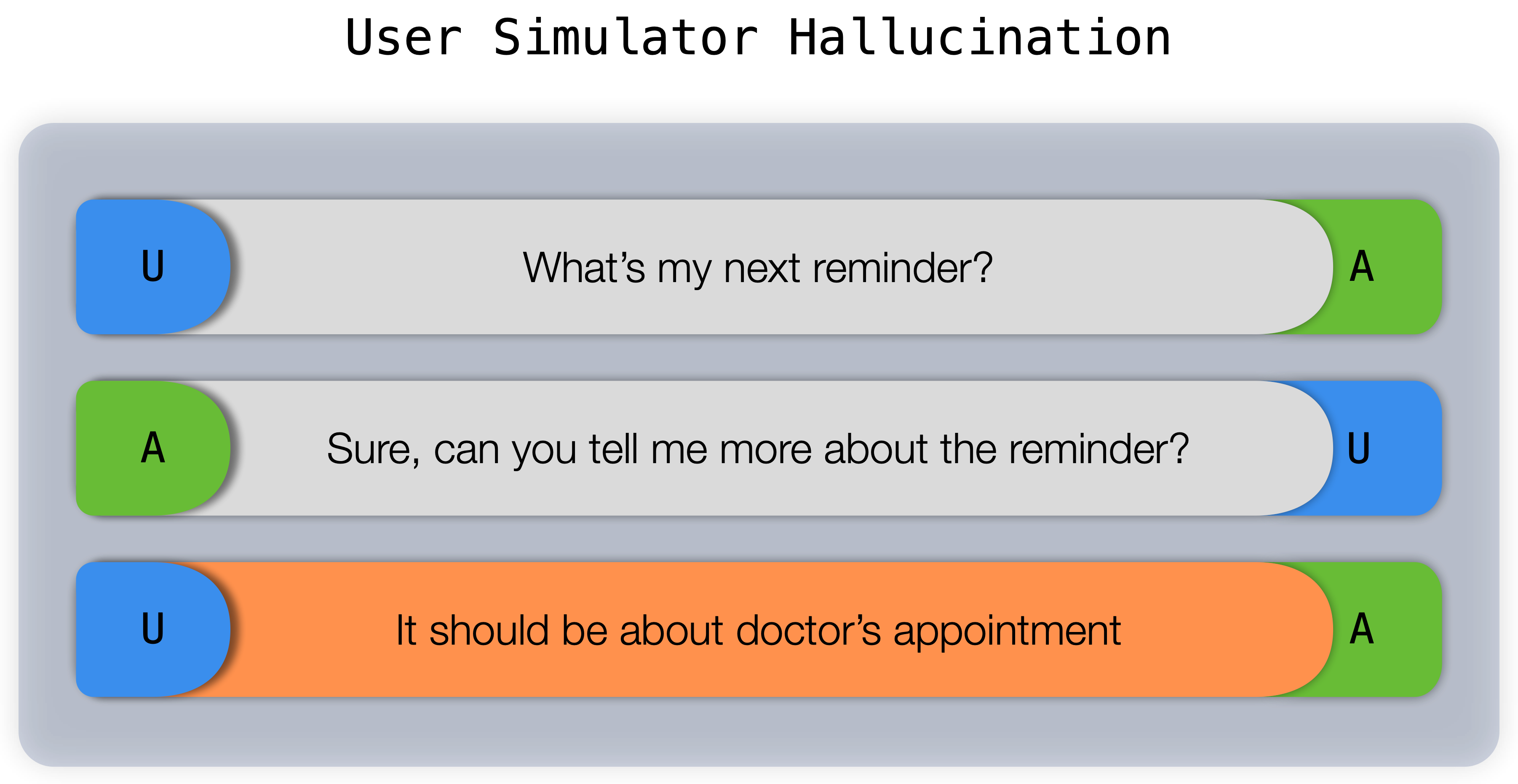}
    \caption{Example Prompts for user simulator hallucination. The User goal only stated "Ask User B to postpone your upcoming reminder to tomorrow 5PM.", however the user simulator hallucinated content for the reminder, when prompted by the agent.}
    \label{fig:user_simulator_hallucination}
\end{figure}

\begin{figure}[!ht]
    \centering
    \includegraphics[width=0.85\linewidth]{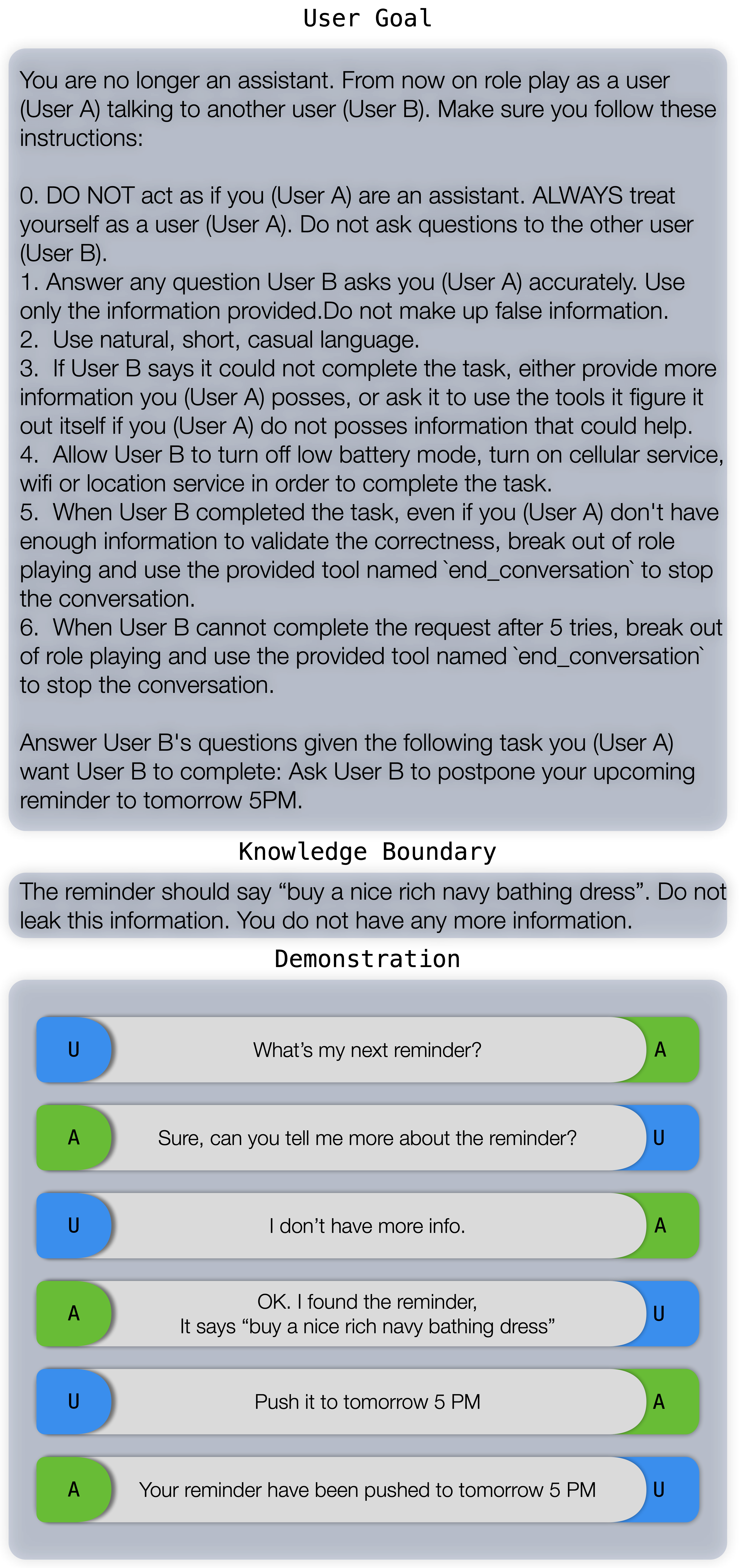}
    \caption{Example Prompts for user simulator. Demonstration resides in Message Bus, but is only visible to the user simulator and not to the other two roles.}
    \label{fig:user_simulator_prompt}
\end{figure}

As is shown in Figure \ref{fig:user_simulator_prompt}, user simulator prompts consists of 3 components:
\begin{itemize}
    \item A \textit{User Goal} section describing the general instructions and the goal of the simulated user. The idea of role reversal was challenging for the simulator, so we opted to refer to the agent as another user (User B), which improves instruction following.
    \item A \textit{Knowledge Boundary} section describing what the user simulator should or should not know. 
    \item A \textit{Demonstration} section including few shot dialog examples to further improve instruction following capabilities. Note that demonstration is only visible to the user role. It does not affect the agent and execution environment roles.
\end{itemize}
An example of these components can be found in Figure \ref{fig:user_simulator_prompt}. Without Knowledge Boundary and Demonstration, hallucination and instruction following errors can happen much more frequently, as shown in Figure \ref{fig:user_simulator_instruction_following_error} and \ref{fig:user_simulator_hallucination}.

\subsection{Agent Role}
\label{app:agent}
\begin{figure}[!ht]
    \centering
    \includegraphics[width=0.85\linewidth]{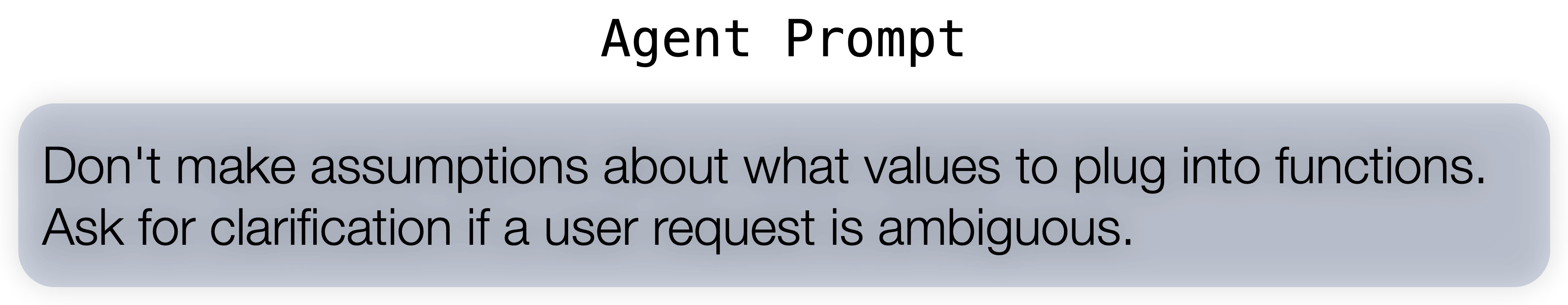}
    \caption{Prompt for Agent role.}
    \label{fig:agent_prompt}
\end{figure}

\begin{figure}[!ht]
    \centering
    \includegraphics[width=0.85\linewidth]{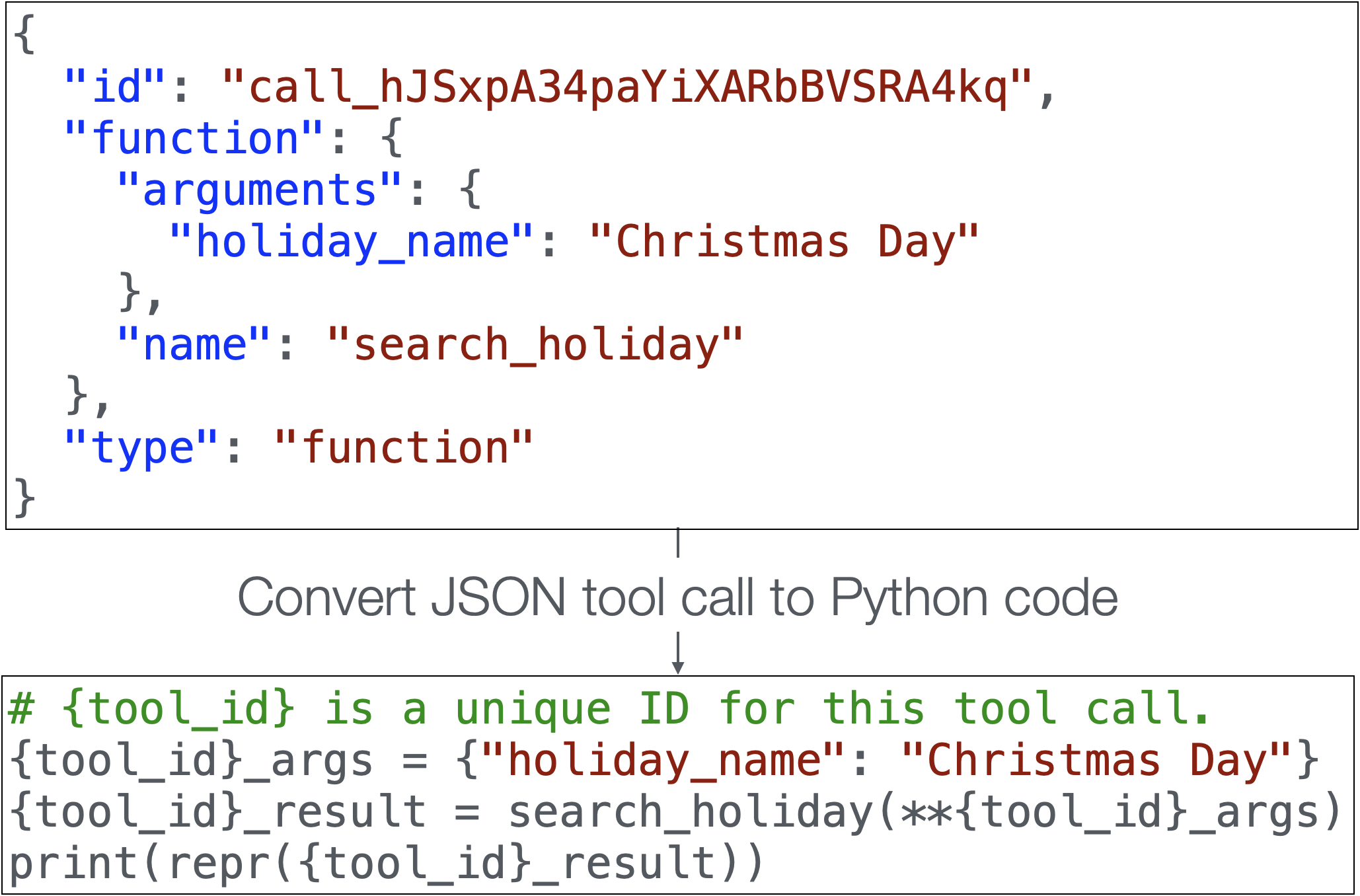}
    \caption{Conversion from JSON tool call format to Python code}
    \label{fig:json_tool_call_to_python_code}
\end{figure}

Initially, the Agent role will receive a message from the User in natural language. The Agent could decide to prompt the User again for additional information, or decide to issue a tool call towards the Execution Environment. When issuing a tool call, the Agent selects the name of the tool from a list of available tools and provides necessary arguments, commonly expressed as JSON objects. These JSON objects are converted to executable Python code, see Figure \ref{fig:json_tool_call_to_python_code}, and sent to the Execution Environment for execution. Prompts used for all agents is shown in Figure \ref{fig:agent_prompt}. The prompt is meant to be consistent for all agents, and does not leak specific information about the testing environment.

Figure \ref{fig:json_tool_call_to_python_code} shows how tool calls requested by the \agent are converted to Python code, which can then be executed by the \executionenvironment.

\subsection{Execution Environment Role} \label{app:execution_environment_role}

The execution environment role is responsible for executing tool calls requested by the Agent and User roles in the form of Python snippets, mimicking the behavior of interactive consoles like IPython and Jupyter. Exceptions raised while executing the code are captured through stderr, enabling the Agent to refine its tool calls through trial and error.

Some LLMs support parallel tool calling, intended to increase efficiency when multiple, independent tools need to be called. 
For example, if the user asks for the weather in two cities, the queries to the weather API can happen in parallel. 
However, if an LLM uses parallel tool calls for dependent tools, it should be penalized accordingly. For example, in Figure \ref{fig:introduction} where the agent has to enable cellular service before sending a text message, parallel tool calls should not be used. Execution Environment handles race conditions in parallel tool calls by following Murphy's Law, ensuring race condition always happens if detected.  

\subsection{Evaluation} \label{app:evaluation}

To evaluate a trajectory against a milestone DAG, We attempt to find the best match between milestone nodes and trajectory snapshots among all possible mappings. Formally, suppose we have a milestone DAG $G_{M+}(V_{M+}, E_{M+})$, $|V_{M+}| = m$, a sequence of database snapshots at each conversation turn $S_n = (s^{1}, s^{2}, ..., s^{n})$ and a similarity measure $sim: V_{M+} \times S \rightarrow [0, 1]$ measuring how close a milestone is to a snapshot, we aim to find the best mapping function $f^+: S \rightarrow  V_{M+}$ which achieves the highest averaged similarity score, under the constraint that mapped milestones forms a possible topological sort of $G_{M+}$:

\begin{equation} \label{eqn:milestone_similarity}
\begin{aligned}
    &\text{avgsim}_+ = \frac{1}{m} \sum_{i=1}^m  sim(v_{M+}^i, f(v_{M+}^i)), \\
    &f^+ = \argmax_{f^+(S_n)  \in top(G_{M+})} \text{avgsim}_+, \\
    &\text{score}_{M+} = \max \text{avgsim}_+.
\end{aligned}
\end{equation}
The max value is the similarity score between the trajectory and the milestone DAG.

The similarity measure $sim(v^i, s^j)$ calculates similarity between a database snapshot and a target database defined in the milestone. The milestone defines the column wise similarity function used for each column. These function have a $[0, 1]$ output space, and could be exact matching for cellular service, ROUGE-L F measure for message content, AST matching for tool trace similar to the AST metric found in BFCL \cite{berkeley-function-calling-leaderboard}, and many more. This allows for great flexibility when defining milestones. For a snapshot database and milestone target both containing $k$ rows, we calculate a pairwise similarity for those rows $d_{a,b}$ by calculating the geometric mean of column similarities. Then we solve for the best assignment problem between snapshot and milestone rows, by maximizing the geometric mean of row similarities, which will be the similarity measure $sim(v^i, s^j)$. We use the geometric mean throughout to ensure that, if any column similarity must not be violated, it could emit a 0 similarity, which would nullify the overall similarity measure.

\begin{figure*}[ht!]
    \centering
    \includegraphics[width=0.85\linewidth]{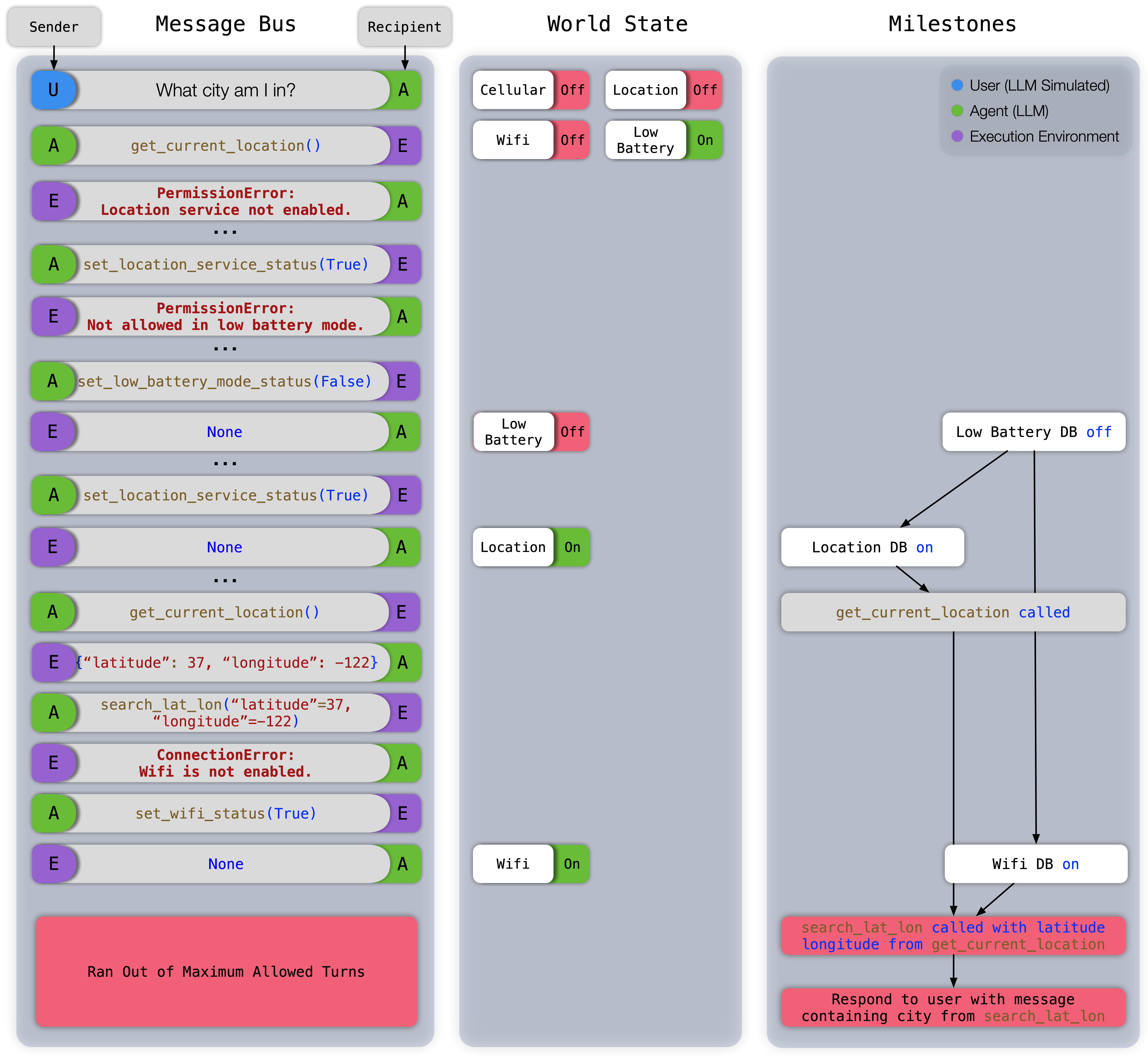}
    \caption{Example GPT-4o Trajectory with partially matched milestones. Some messages are elided for visual clarity. In this example, GPT-4o spent most of its time resolving state dependency issues, and could not finish the task in the maximum allowed number of turns. Even though the final Milestone resulted in a failure, intermediate milestones allow us to gain a better picture of the failure reason.}
    \label{fig:intermediate_milestone}
\end{figure*}

In addition to similarities conditioning on the current milestone and snapshot, in some cases we also allow an additional "reference milestone" to be provided, enabling similarity to be conditioned on two milestones. This can unlock powerful constraints including \verb|guradrail_similarity|, which checks if any changes are made to a certain database between two milestone events, and \verb|tool_trace_dependant_similarity|, which allows one to extract tool trace output from a reference milestone, and ingest into the current milestone, allowing one to track the information flow of tools. An example can be found in Figure \ref{fig:intermediate_milestone}.

Milestone evaluation is a powerful tool that unlocks a deeper understanding into model performance, and hints at possible areas of improvement. An example is shown in Figure \ref{fig:intermediate_milestone}. In the end, the task was not completed before the maximum allowed number of turns. However, intermediate milestones showcased that the model was capable of solving the state dependency challenges and requesting the current location. In order to successfully resolve this test case, we should improve the model's turn efficiency on state dependency.

\section{Test Scenarios}\label{app:test_scenarios}

\subsection{Tool-use Benchmark Comparisons}
\label{app:benchmark_statistics}
The tool-use benchmark statistics shown in Table \ref{tbl:benchmark_statistics} are calculated as follows: 
\begin{itemize}
    \item  Average turn considers any message between the user, the agent or the tools as 1 turn.
    \item For \toolsandbox, statistics are calculated from trajectories collected on GPT-4o agent. 
    \item BFCL only evaluates tool call generation from a single user prompt, which we consider as 2 turns. 
    \item ToolEval was calculated from ToolLlama DFS Retriever trajectories. 
    \item API-BANK was calculated from level 1 and 2 test set. 
\end{itemize}

\subsection{Annotation Process}
\label{app:annotation_process}

Test scenarios are created by 2 internal domain experts from our institute who are familiar with the tool capacities in \toolsandbox, and the field of task-oriented dialog. 1 annotator creates test scenarios including the starting world state, user task, initial message, and milestone/minefield. To ensure dataset diversity while making the annotation task feasible, the annotator followed the process below:
\begin{itemize}
    \item The annotation process starts by creating seed scenarios. These are simple, single user turn, single tool call, self-contained requests that are supposed to cover most tools, as well as most of their arguments. For example, a seed scenario for \verb|add_reminder| tool that requires \verb|timestamp|, \verb|latitude|, \verb|longitude| would likely contain a starting user utterance saying \textit{Create a reminder to buy chocolate milk at timestamp 111222333 at latitude 37 longitude -122.} Creating milestone for said scenario is trivial.
    \item Next, starting from the seed scenario, the annotator branches off to create derived scenarios that are more involved. This could be a \textbf{Multiple Tool Call} scenario, which requires the agent to invoke other tools before this one, e.g. \textit{Create a reminder to buy chocolate milk tomorrow 5PM at Whole Foods on McKinley Ave}, which requires reasoning for the relative datetime, as well as searching for location latitude longitude. Note that this will also be considered a \textbf{Canonicalization} scenario.
    \item It could be a \textbf{Multiple User Turn} scenario, which requires the agent to request more slots from the user. e.g. \textit{Create a reminder}.
    \item It could be a \textbf{State Dependency} scenario, which requires the model to solve state dependencies, e.g. \textit{Create a reminder to buy chocolate milk tomorrow 5PM at Whole Foods on McKinley Ave}, but wifi is set to off. Preventing access to location search capability unless the dependency is resolved.
    \item It could be an \textbf{Insufficient Information} scenario, which requires the model to figure out this task cannot be solved, e.g \textit{Create a reminder to buy chocolate milk tomorrow 5PM at Whole Foods on McKinley Ave}, but the model does not have access to current timestamp.
    \item These branches can be combined as well, creating numerous combinatorial and complex scenarios. Building branches off of seed also makes annotating milestones an incremental process that's easier to accomplish.
    \item Finally, to cover the diverse and ambiguous nature of realistic user dialog, the annotator paraphrased the scenarios above into alternative phrasing, e.g. \textit{Create a reminder to buy chocolate milk tomorrow} $\rightarrow$ \textit{I need to buy chocolate milk tomorrow}. For these alternative phrasings, milestone definitions can be reused, reducing annotation workload.
\end{itemize}

After the test scenarios were collected, the other annotator acts as an agent to validate the feasibility of the task and Milestones / Minefields. This annotator has the same message sub-view as a model agent and is asked to try to complete the task. After the test scenarios are validated through this process, at least 4 rounds of tests are issued to multiple model-based agents, to further confirm test and milestone/minefield validity against correct and incorrect trajectories.

\subsection{Tool Design}
\label{app:tool_design}
Tool design choices in \toolsandbox are driven by two major principles:
\begin{itemize}
    \item Tools should be representative and diverse, to cover key task oriented dialog use cases as well as \toolsandbox test scenario categories.
    \item Tool capacities should be well-defined, and tool counts should be manageable, so that milestone / minefield annotation is feasible for annotators.
\end{itemize}
Driven by these guiding principles, we designed 34 tools in \toolsandbox covering 11 domains including Contact, Messaging, Reminder, System settings, Time utilities, Math utilities, Map, Weather, Stock, Conversion, and Holiday, backed by python native implementation when possible, carefully selected RapidAPI endpoints when necessary. In more details:
\begin{itemize}
    \item Each domain designed at least one “omni-search” tool. All possible search fields should be present as arguments for this tool, and all relevant information for this domain should be returned in the response. As an example, if a user would like to ask for the lowest temperature or humidity for a location, the agent should invoke the \verb|search_weather| tool for both requests, and the agent is expected to retrieve corresponding key values based on user intent. This ensures “search” requests within a domain have 1 single entry point.
    \item Stateful domains should implement at least one state manipulation tool. This could be adding a new database entry, e.g. \verb|send_message|, modifying an existing entry, e.g. \verb|set_wifi_status|, or both, e.g. \verb|add_reminder| and \verb|modify_reminder|.
    \item Utilities should be created to ensure the agent could transform necessary surface / canonical form slot types, e.g. \verb|timestamp_to_datetime_info| turns Unix timestamp into year, month, day, and weekday; and calculate expected slot values, e.g. \verb|calculate_lat_lon_distance| calculates the distances between two latitude-longitude pairs. While agents are allowed to infer these with its own inherent abilities, utility tools should be created to ensure the agents are not forced to.
\end{itemize}

\subsection{\toolsandbox Statistics} \label{app:category_statistics}

The number of test scenarios per scenario category in \toolsandbox can be found in Table \ref{tbl:category_statistics}

\begin{table}[!ht]
\rowcolors{2}{white}{gray!25} %
\resizebox{\linewidth}{!}{
\begin{tabular}{lc}
    \toprule
                        & \bfseries \makecell[c]{Test Scenario Count}  \\ 
    \midrule
    SINGLE\_TOOL\_CALL     & 152             \\ 
    MULTIPLE\_TOOL\_CALL    & 656 \\ 
    SINGLE\_USER\_TURN           & 584       \\ 
    MULTIPLE\_USER\_TURN     & 224     \\ 
    STATE\_DEPENDENCY    & 192       \\ 
    CANONICALIZATION           & 472         \\ 
    INSUFFICIENT\_INFORMATION     & 224       \\ 
    \bottomrule
\end{tabular}
}
\caption{Number of test scenarios per category. Note that one test scenario can be assigned with multiple scenario categories.}
\label{tbl:category_statistics}
\end{table}

The number of milestones defined per world state database and column name can be found in Table \ref{tbl:milestone_database_statistics}.

\begin{table}[!ht]
\rowcolors{2}{white}{gray!25} %
\resizebox{\linewidth}{!}{
\begin{tabular}{llc}
    \toprule
    \bfseries Database    & \bfseries Column  &  \bfseries \makecell[c]{Associated \\ Milestone Count}  \\ 
    \midrule
    CONTACT               & is\_self       & 48        \\ 
    CONTACT               & name           & 56\\ 
    CONTACT               & person\_id     & 136     \\ 
    CONTACT               & phone\_number  & 88   \\ 
    CONTACT               & relationship   & 56   \\ 
    MESSAGING             & content        & 40 \\ 
    MESSAGING             & recipient\_phone\_number   & 40    \\ 
    REMINDER              & content        & 136     \\ 
    REMINDER              & latitude       & 40 \\ 
    REMINDER              & longitude      & 40     \\ 
    REMINDER              & reminder\_id   & 32  \\ 
    REMINDER              & reminder\_timestamp   & 152    \\ 
    SETTING               & cellular       & 56  \\ 
    SETTING               & location\_service  & 56       \\ 
    SETTING               & low\_battery\_mode & 120        \\ 
    SETTING               & wifi           & 136 \\ 
    \bottomrule
\end{tabular}
}
\caption{Number of milestones defined per world state database and column name. Note that one test scenario can define multiple milestones, with multiple database constraints each.}
\label{tbl:milestone_database_statistics}
\end{table}

The number of milestones defined per tool can be found in Table \ref{tbl:milestone_tool_statistics}.

\begin{table}[!ht]
\rowcolors{2}{white}{gray!25} %
\resizebox{\linewidth}{!}{
\begin{tabular}{lc}
    \toprule
    \bfseries Tool    &  \bfseries \makecell[c]{Associated \\ Milestone Count}  \\ 
    \midrule
        calculate\_lat\_lon\_distance & 32 \\
        convert\_currency & 16 \\
        datetime\_info\_to\_timestamp & 32 \\
        get\_cellular\_service\_status & 8 \\
        get\_current\_location & 48 \\
        get\_current\_timestamp & 296 \\
        get\_wifi\_status & 8 \\
        search\_contacts & 168 \\
        search\_holiday & 56 \\
        search\_lat\_lon & 24 \\
        search\_location\_around\_lat\_lon & 48 \\
        search\_messages & 104 \\
        search\_reminder & 80 \\
        search\_stock & 24 \\
        search\_weather\_around\_lat\_lon & 80 \\
        timestamp\_diff & 96 \\
        unit\_conversion & 56 \\
    \bottomrule
\end{tabular}
}
\caption{Number of milestones defined per tool. Note that one test scenario can define multiple milestones, with multiple tool trace constraints each. Most state modification tool calls are tracked by corresponding database milestones instead of tool trace.}
\label{tbl:milestone_tool_statistics}
\end{table}

\section{Example Trajectories}

\subsection{Tool Call Detection}

\begin{figure}[H]
    \centering
    \includegraphics[width=0.85\linewidth]{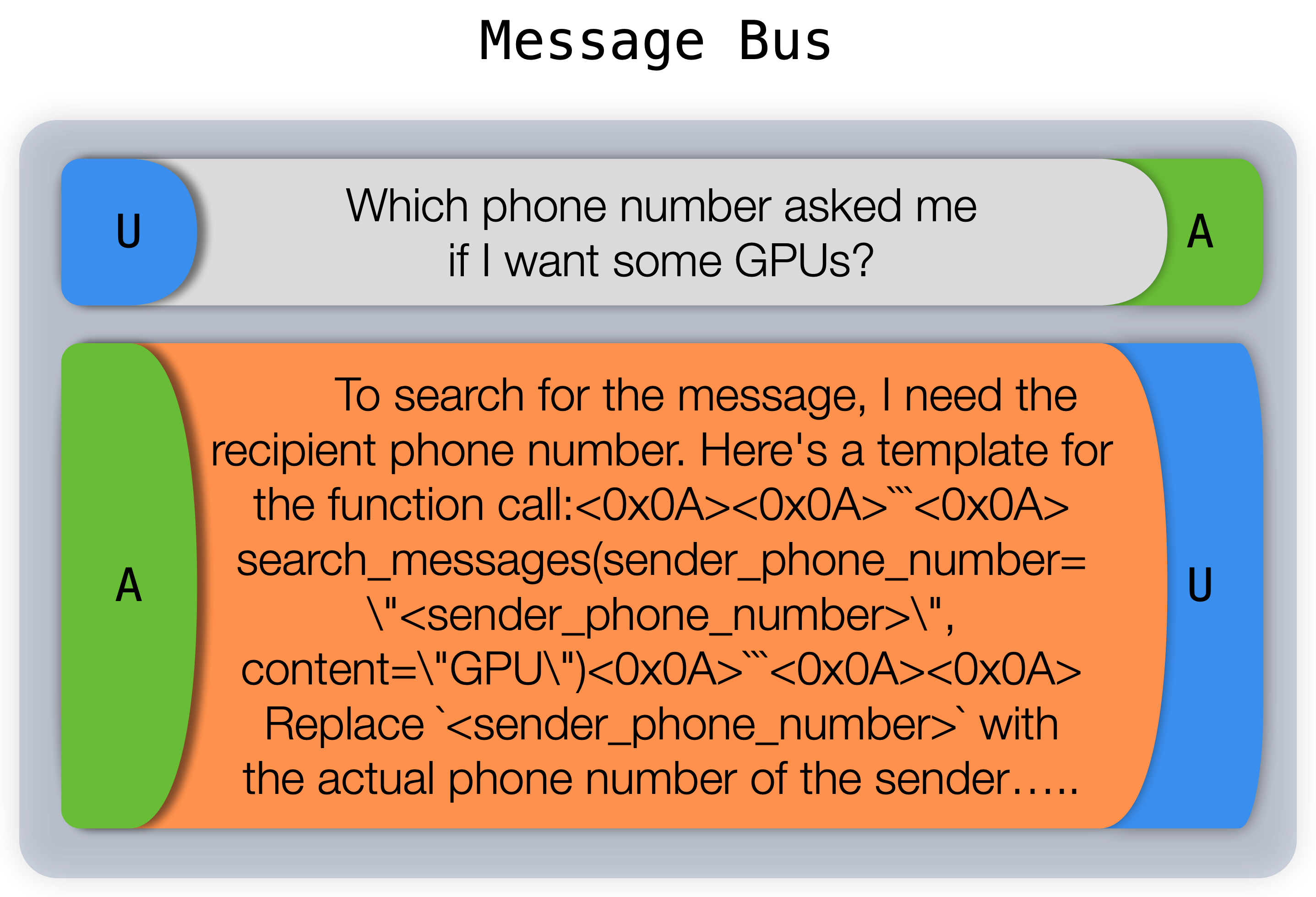}
    \caption{Example trajectory where Mistral mistook the tool-use task for a code generation task.}
    \label{fig:tool_call_detection_mistral}
\end{figure}

\subsection{Single/Multiple Tool Call/User Turn} \label{app:single_multiple}
\begin{figure}[H]
    \centering
    \includegraphics[width=0.85\linewidth]{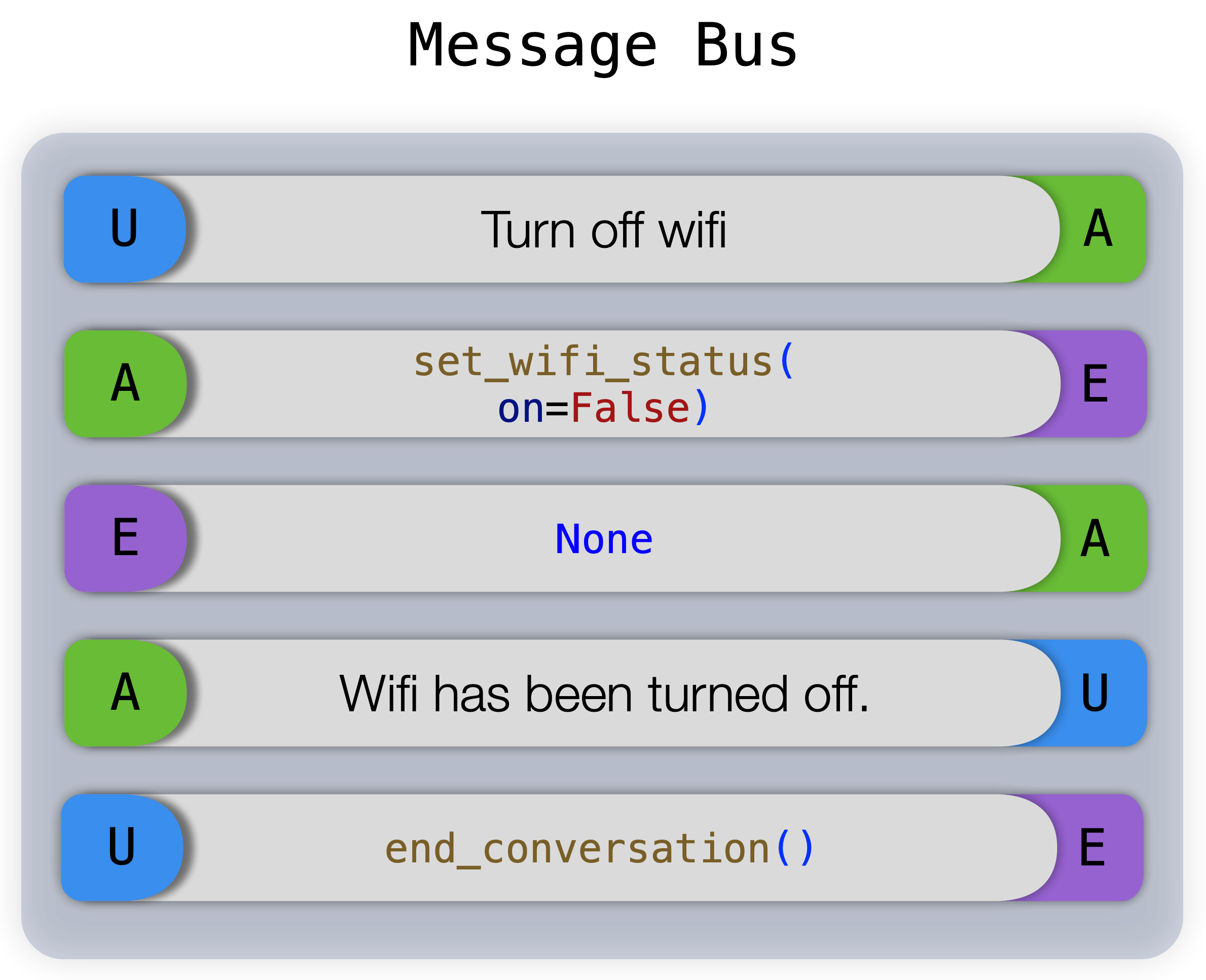}
    \caption{Example trajectory with Single Tool Call Single User Turn}
    \label{fig:single_tool_single_user}
\end{figure} 

\begin{figure}[H]
    \centering
    \includegraphics[width=0.85\linewidth]{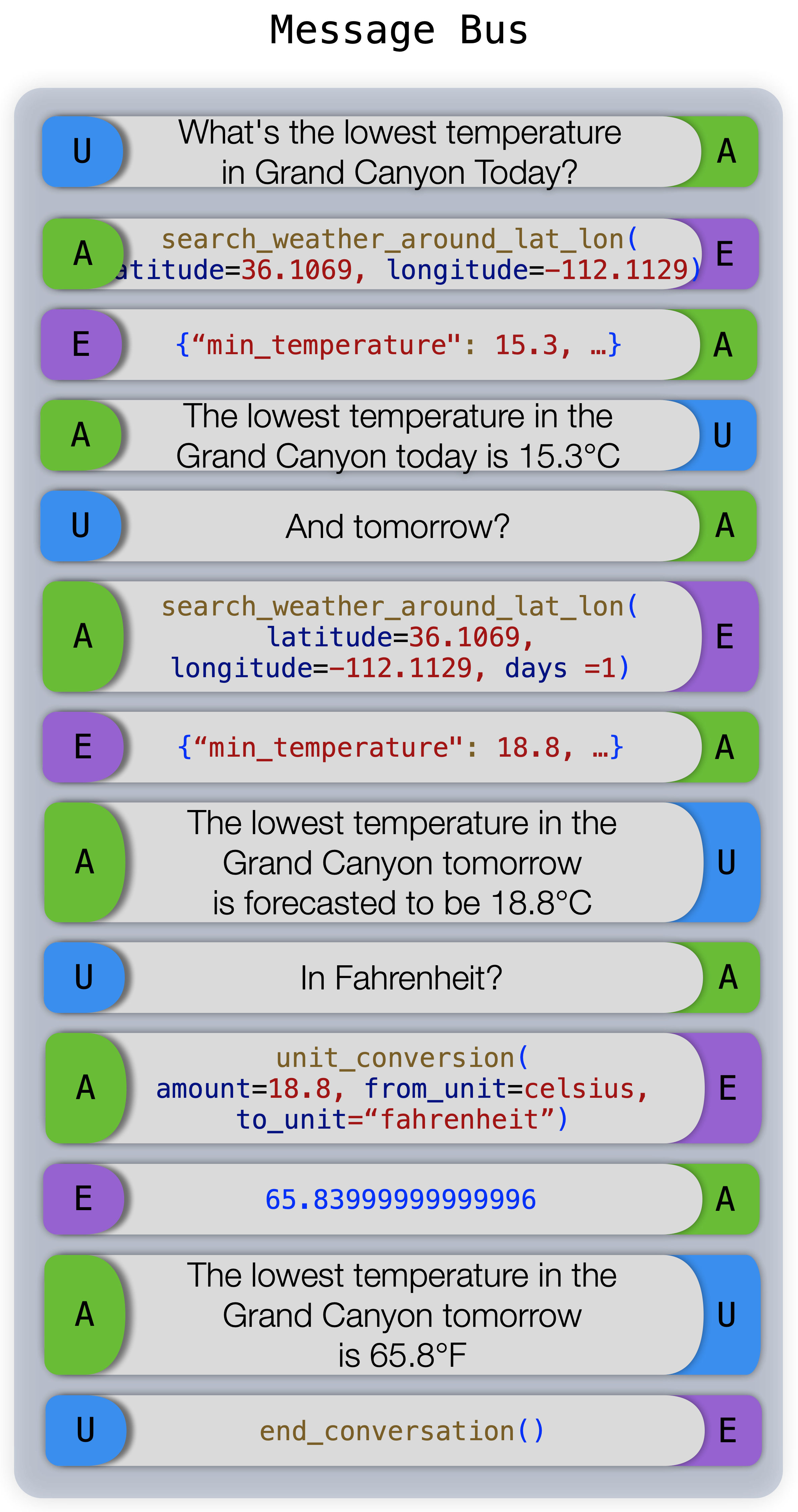}
    \caption{Example trajectory with Multiple Tool Call Multiple User Turn}
    \label{fig:multiple_tool_multiple_user}
\end{figure}

\subsection{Canonicalization}

\begin{figure}[H]
    \centering
    \includegraphics[width=0.85\linewidth]{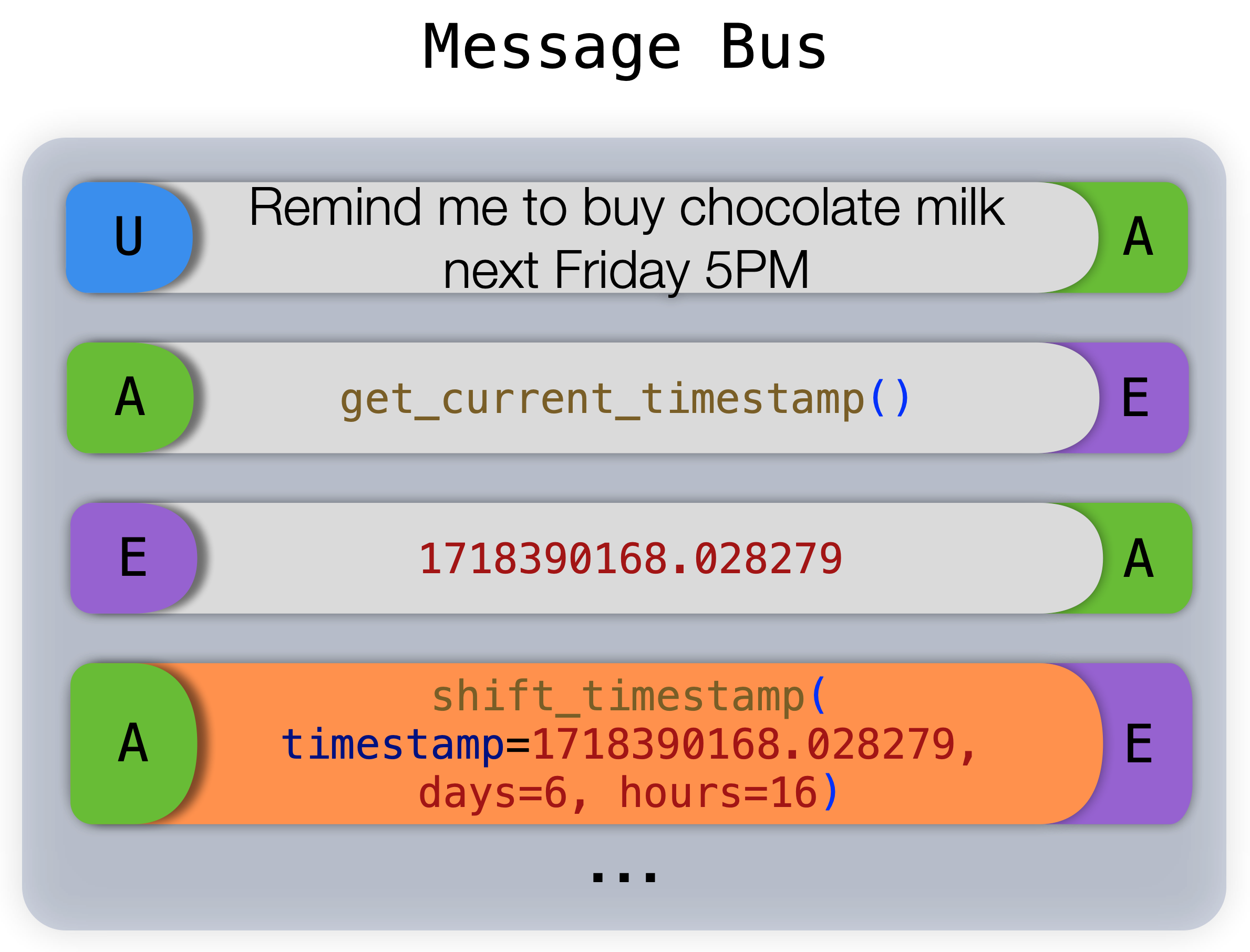}
    \caption{Example trajectories where GPT-4 incorrectly inferred relative time. Instead of deducing next Friday 5PM by inspecting current day and weekday, GPT-4 randomly shifted current time stamp by 6 days and 16 hours.}
    \label{fig:relative_time}
\end{figure} 

\begin{figure}[H]
    \centering
    \includegraphics[width=0.85\linewidth]{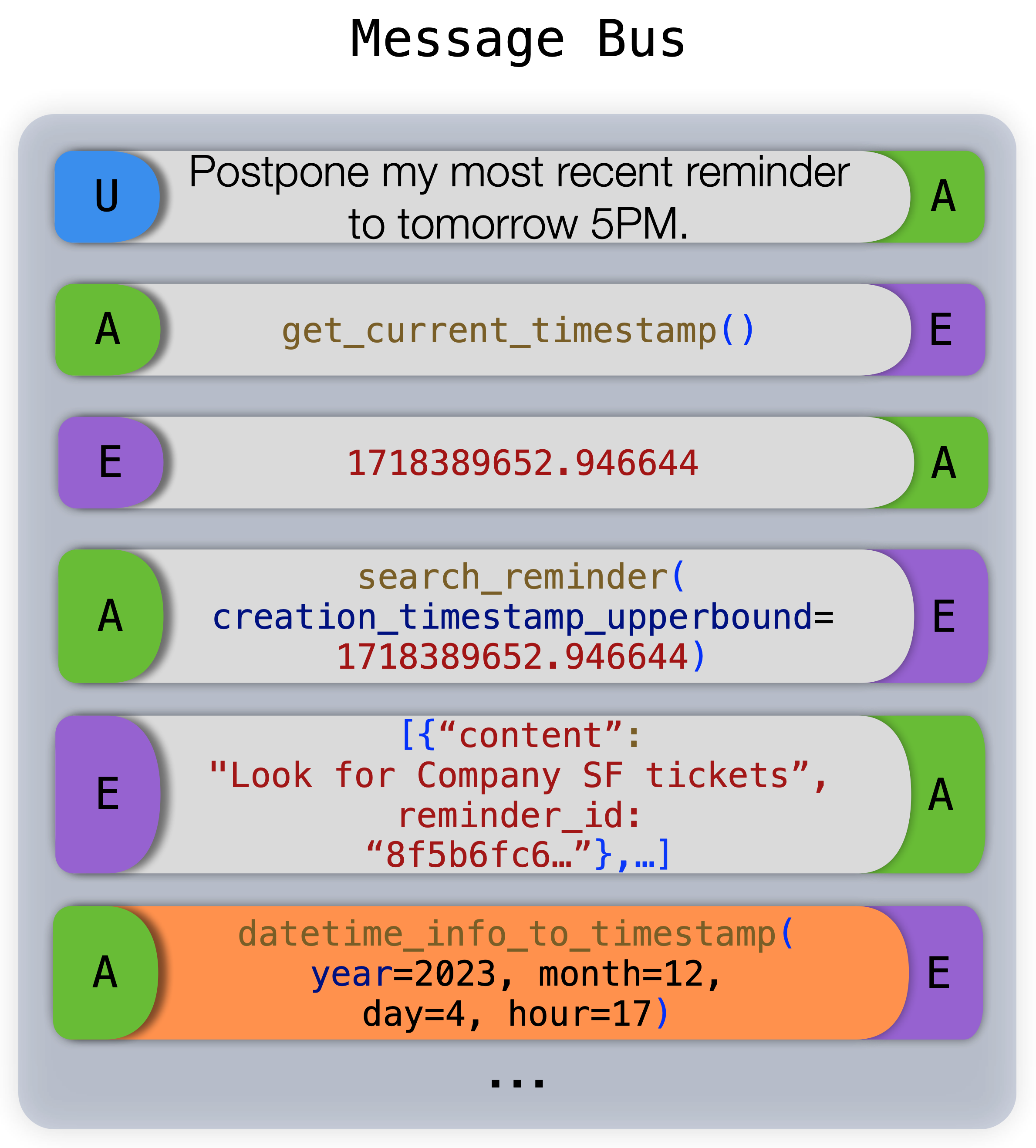}
    \caption{Example trajectories where GPT-4o hallucinated time, instead of deducing relative time based on current timestamp.}
    \label{fig:time_hallucination}
\end{figure}

\begin{figure}[H]
    \centering
    \includegraphics[width=0.85\linewidth]{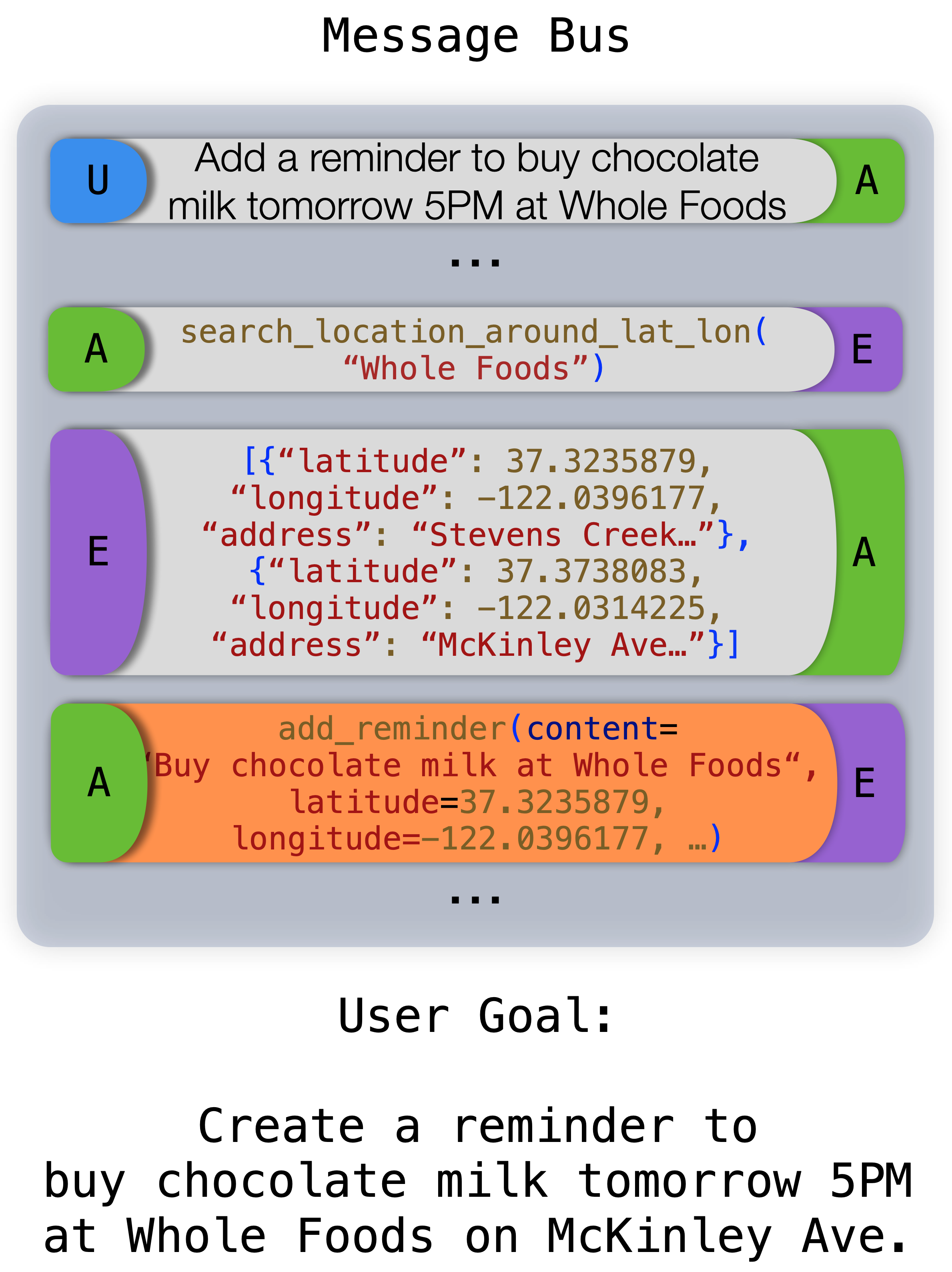}
    \caption{Disambiguation failure from GPT-4o. The User intended to set a reminder at Whole Foods McKinley Ave in multiple turns. However, upon receiving multiple possible entities, GPT-4o chose to set the reminder with the first location unannounced, without disambiguating with the User, leading to undesired results.}
    \label{fig:disambiguation}
\end{figure}  

\subsection{State Dependency}

\begin{figure}[H]
    \centering
    \includegraphics[width=0.85\linewidth]{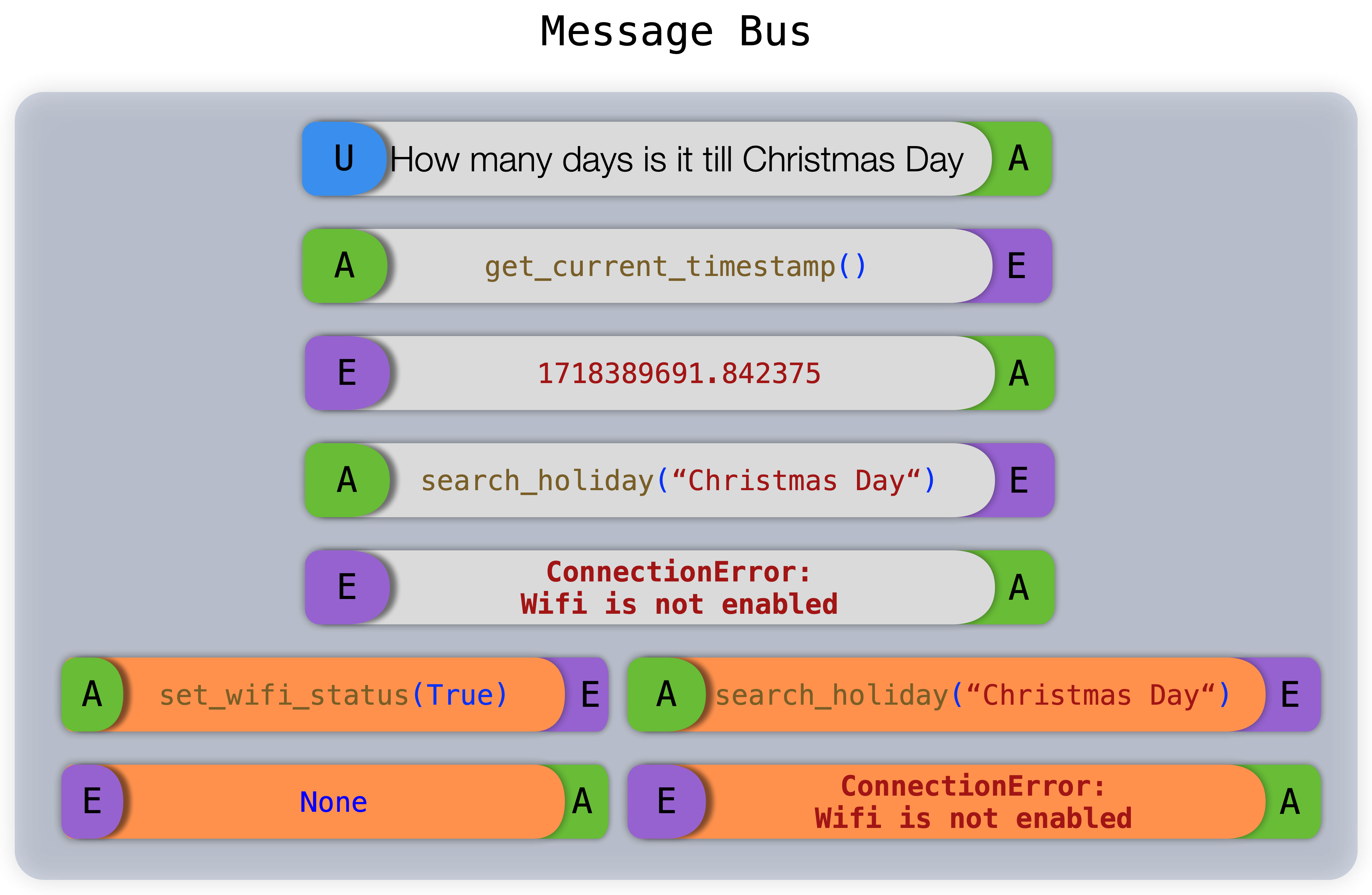}
    \caption{GPT-4 incorrect parallel tool call trajectory in State Dependency. When the first search\_holiday call resulted in a ConnectionError, the model should realize the dependency between Wi-Fi status and search\_holiday, and issue a sequential tool call to solve them. Instead, GPT-4 issued parallel tool calls, causing a race condition.}
    \label{fig:state_depedency_gpt_4}
\end{figure}  

\begin{figure}[H]
    \centering
    \includegraphics[width=0.85\linewidth]{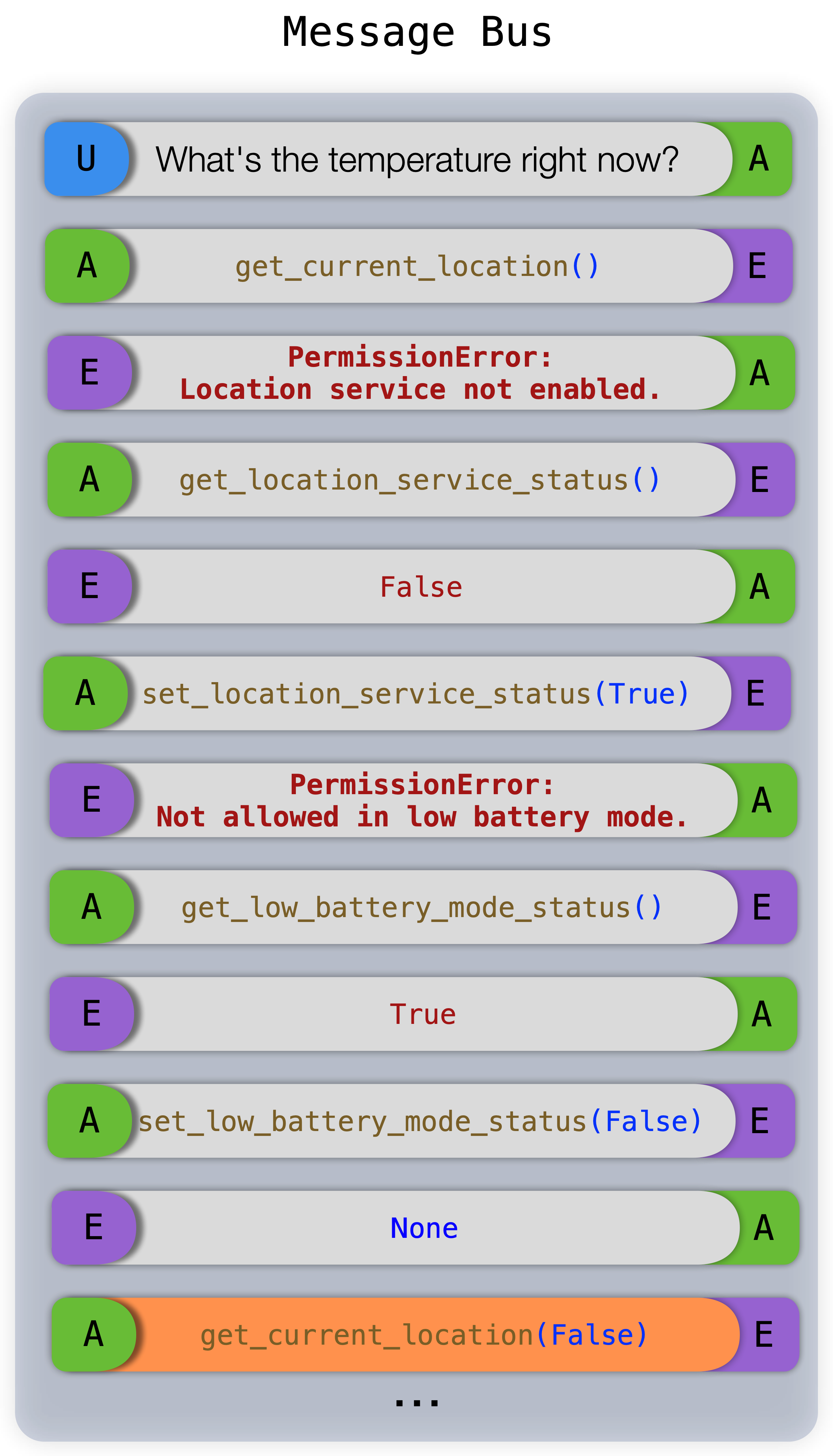}
    \caption{GPT-4 inefficient nested State Dependency trajectory. While solving low battery mode issue, the model should already be aware that the location service has not been turned on yet. Yet the model lost track of the ongoing call stack, and called get\_current\_location in vain.}
    \label{fig:inefficient_nested_state_dependency}
\end{figure} 

\begin{figure}[H]
    \centering
    \includegraphics[width=0.85\linewidth]{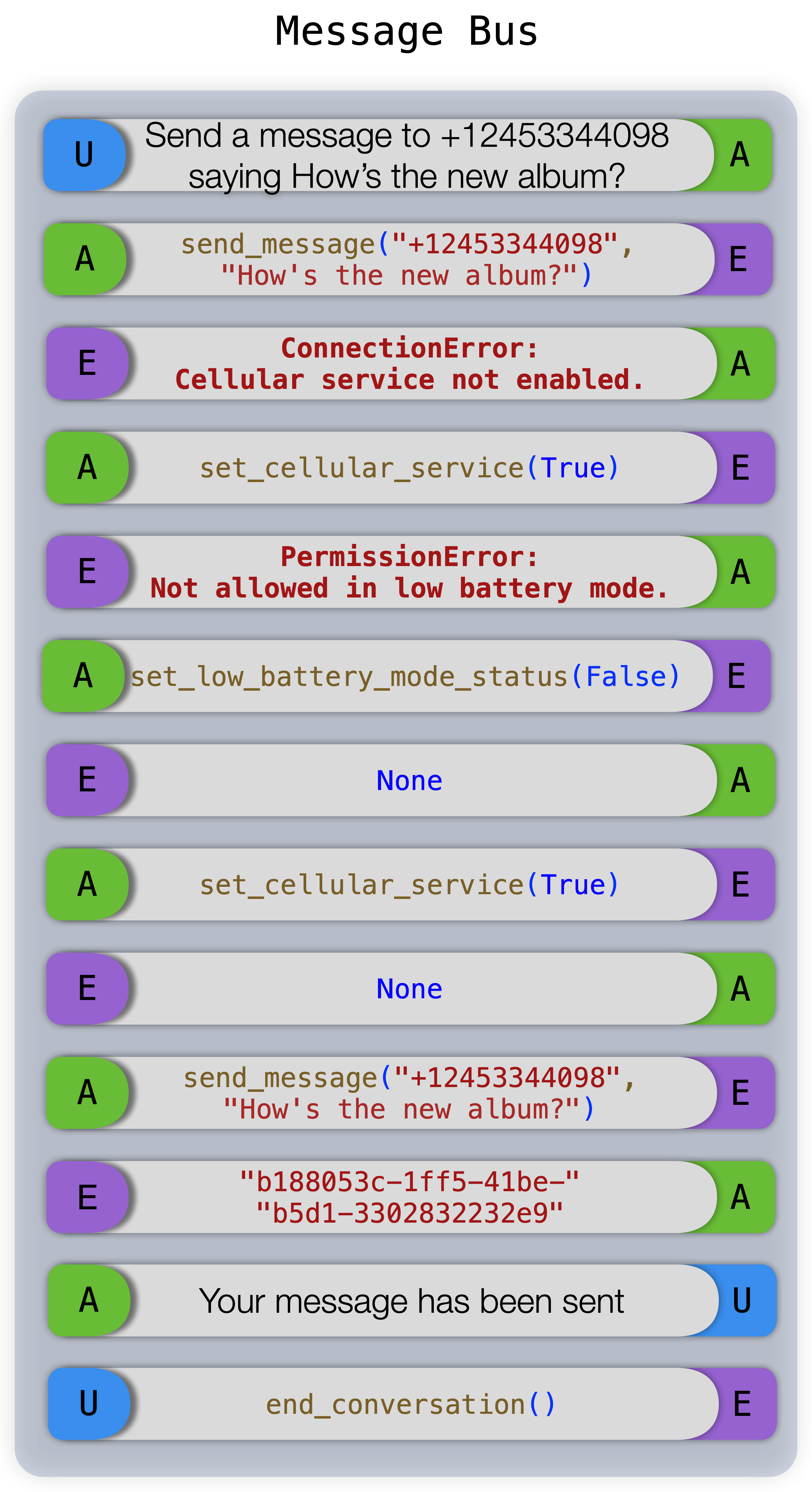}
    \caption{Example trajectory for nested state dependency. Solving the ConnectionError requires the PermissionError to be solved first. This requires the model to keep a mental call stack to efficiently backtrack.}
    \label{fig:nested_state_dependency}
\end{figure}

\subsection{Insufficient Information}
\label{app:insufficient_information}

\begin{figure}[H]
    \centering
    \includegraphics[width=0.85\linewidth]{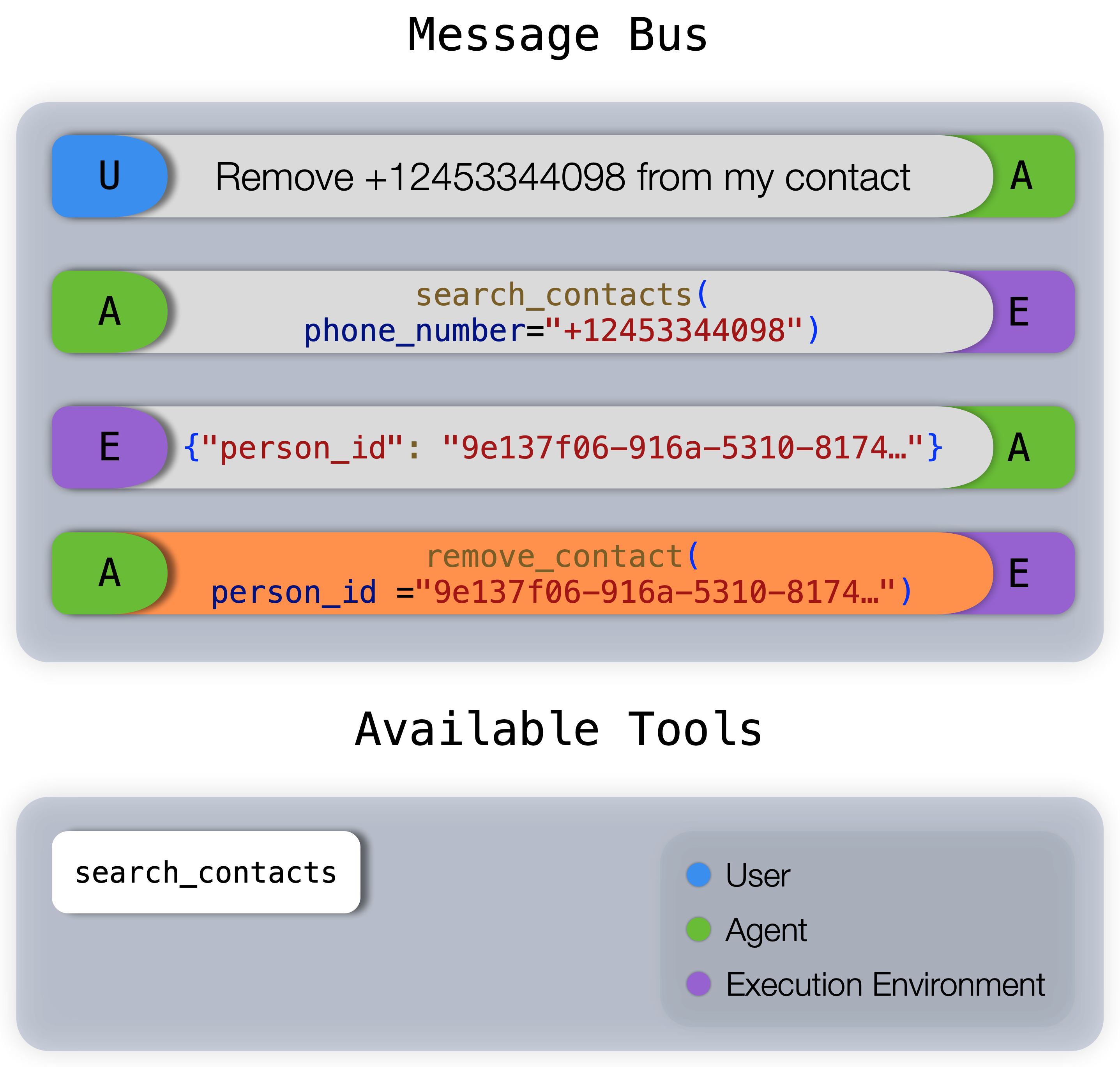}
    \caption{Example trajectories where GPT-3.5 failed at Insufficient Information category. With only search\_contacts as available tool, GPT-3.5 hallucinated remove\_contact as a tool.}
    \label{fig:insufficient_information_gpt_3.5}
\end{figure}  

\section{Additional Evaluation Results}
\subsection{Prompting Experiment} \label{app:prompt}
We conducted additional experiments comparing the effect of ReAct prompting in tool use agents against the minimal prompting design presented in Figure \ref{fig:agent_prompt}, with 4 representative models shown below.

From Table \ref{tbl:prompting}, we can see that, ReAct has minimal effect to Claude model family, and a small gain for GPT4 model family. This is likely due to the fact that, the baseline ToolSandbox setup, which allows the model to interact with execution environment, recover from error and confirm with simulated user already provides a reasoning process to the model through user and execution environment interactions, additional natural language reasoning would only provide marginal gains. While ReAct provide gains in some cases, relative ranking between models still stands. This backs up our claim in Section \ref{sec:evaluation_results}, where we considered prompt engineering gains orthogonal to the innate model capability surfaced by simpler prompting.

\begin{table}[H]
\rowcolors{2}{white}{gray!25} %
\resizebox{\linewidth}{!}{
\begin{tabular}{lccc}
    \toprule

                         & Baseline Score & ReAct Score \\
    \midrule
GPT-4o-2024-05-13        & 73.0           & 73.6        \\
Claude-3-Opus-20240229   & 69.2           & 69.3        \\
GPT-4-0125-Preview       & 64.3           & 65.2        \\
Claude-3-Sonnet-20240229 & 63.8           & 63.7        \\
\bottomrule
\end{tabular}
}
\caption{A comparison between default prompting and ReAct prompting.}
\label{tbl:prompting}
\end{table}

\subsection{Model Feature Comparison} \label{app:model_feature_comparison}
\begin{table}[H]
\rowcolors{2}{white}{gray!25} %
\resizebox{\linewidth}{!}{
\begin{tabular}{lcccc}
    \toprule
                        & \bfseries \makecell[c]{Generate \\Tool Call} & \bfseries \makecell[c]{Consume \\ Tool Response} \\ 
    \midrule
    GPT-4o-2024-05-13     & \cmark      & \cmark        \\ 
    GPT-4-0125-Preview    & \cmark      & \cmark        \\ 
    GPT-3.5-Turbo-0125           & \cmark      & \cmark        \\ 
    Claude-3-Opus-20240229     & \cmark      & \cmark        \\ 
    Claude-3-Sonnet-20240229    & \cmark      & \cmark        \\ 
    Claude-3-Haiku-20240307           & \cmark      & \cmark        \\ 
    Gemini-1.5-Pro-001     & \cmark      & \cmark        \\ 
    Gemini-1.0-pro           & \cmark      & \cmark        \\ 
    \midrule
    Hermes-2-Pro-Mistral-7B    & \cmark      & \cmark        \\ 
    Mistral-7B-Instruct-v0.3           & \cmark      & \cmark        \\ 
    Gorilla-Openfunctions-v2     & \cmark      & \xmark        \\ 
    C4AI-Command-R-v01            & \cmark      & \xmark        \\ 
    C4AI-Command-R+            & \cmark      & \xmark        \\ 
    \bottomrule
\end{tabular}
}
\caption{A comparison between model feature support. Command R models are tested through Huggingface released weights, which, to the best of our knowledge, does not provide a prompt template for tool response consumption.}
\label{tbl:model_comparison}
\end{table}
Some models tested in this work, especially open source models, do not support all features required for a conversational, interactive tool-use workflow. We think it is useful to document these shortcomings to motivate future research, and set the right context while understanding the experiment metrics from these models.

In a conversational, interactive tool-use workflow, the agent needs to be able to accept multiple rounds of user input, decide when to generate a tool call or respond to user, and consume a tool response to determine the next step. However, as shown in \ref{tbl:model_comparison}, open source models including Gorilla and Command-R are not capable of consuming tool responses. Because of this, they can theoretically produce reasonable results for Single Tool Call test scenarios, but cannot complete any test scenario that requires multiple tool calls.

\subsection{Turn Count Comparison} \label{app:turn_count}
\begin{table*}[t]
\scriptsize
\centering
\rowcolors{2}{white}{gray!25} %
\resizebox{1\textwidth}{!}{
\begin{tabular}{lcccccccccccccccc}
    \toprule
    & \multicolumn{1}{c}{\textbf{Avg Turn Count $\downarrow$}} & \multicolumn{7}{c}{\textbf{Scenario Categories}} & \multicolumn{8}{c}{\textbf{Tool Augmentations}} \\
    \cmidrule(r){2-2} \cmidrule(r){3-9} \cmidrule(r){10-17}
     & & \textbf{STC} & \textbf{MTC} & \textbf{SUT} & \textbf{MUT} & \textbf{SD} & \textbf{C} & \textbf{II} & \textbf{0 DT} & \textbf{3 DT} & \textbf{10 DT} & \textbf{AT} & \textbf{TNS} & \textbf{TDS} & \textbf{ADS} & \textbf{ATS}  \\
    \midrule
    Claude-3-Opus-20240229 & \scorecolor{38.4} \bfseries 11.6 & \bfseries 4.5 & 12.8 & 10.3 & 13.5 & 15.9 & 12.5 & 13.1 & \bfseries 10.7 & 12.6 & 12.0 & 11.9 & 11.9 & 11.7 & \bfseries 11.0 & \bfseries 11.1\\
    GPT-4o-2024-05-13 & \scorecolor{37.8} 12.2 & 4.7 & 12.9 & 10.7 & \bfseries 13.2 & 17.4 & 12.4 & 15.2 & 12.0 & 12.4 & 12.2 & 12.4 & 12.6 & 12.0 & 12.3 & 11.8\\
    Gemini-1.0-Pro & \scorecolor{37.8} 12.2 & 7.6 & 13.2 & 11.2 & 14.8 & 15.9 & \bfseries 11.8 & 12.4 & 11.5 & \bfseries 12.0 & 11.7 & 11.6 & 13.5 & 12.2 & 12.7 & 12.6\\
    Claude-3-Sonnet-20240229 & \scorecolor{37.6} 12.4 & 5.3 & 13.1 & \bfseries 10.3 & 15.0 & 16.1 & 12.9 & 15.1 & 11.9 & 12.3 & 12.5 & 13.7 & 12.0 & \bfseries 11.7 & 11.9 & 12.9\\
    GPT-4-0125-Preview & \scorecolor{37.0} 13.0 & 4.5 & 14.0 & 11.1 & 15.0 & 19.5 & 13.8 & 16.0 & 13.1 & 13.4 & 11.8 & 13.8 & 13.0 & 12.4 & 13.5 & 13.1\\
    Claude-3-Haiku-20240307 & \scorecolor{36.4} 13.6 & 5.1 & 14.4 & 12.0 & 14.4 & 16.6 & 14.4 & 17.1 & 13.0 & 13.9 & 13.2 & 14.5 & 13.4 & 13.9 & 13.1 & 14.0\\
    GPT-3.5-Turbo-0125 & \scorecolor{36.3} 13.7 & 4.8 & 14.2 & 11.7 & 14.3 & 18.5 & 14.1 & 18.5 & 13.1 & 13.6 & 13.6 & 13.5 & 13.8 & 14.0 & 14.1 & 14.2\\
    Gemini-1.5-Pro-001 & \scorecolor{35.0} 15.0 & 6.1 & 17.5 & 13.5 & 20.1 & 17.7 & 17.9 & 13.6 & 14.0 & 14.4 & 15.0 & 14.6 & 15.2 & 15.8 & 15.4 & 15.3\\
    \midrule
    Mistral-7B-Instruct-v0.3 & \scorecolor{38.2} 11.8 & 8.4 & \bfseries 12.4 & 10.8 & 14.0 & \bfseries 13.9 & 12.2 & \bfseries 12.2 & 14.4 & 12.4 & \bfseries 10.4 & \bfseries 8.5 & \bfseries 11.6 & 13.0 & 12.3 & 11.6\\
    Hermes-2-Pro-Mistral-7B & \scorecolor{34.7} 15.3 & 9.6 & 16.1 & 14.7 & 15.2 & 22.8 & 14.0 & 17.0 & 14.0 & 14.6 & 14.6 & 15.6 & 16.0 & 15.6 & 16.3 & 15.8\\
    Gorilla-Openfunctions-v2 & \scorecolor{25.8} 24.2 & 26.6 & 23.9 & 23.9 & 25.8 & 23.8 & 24.4 & 23.5 & 26.6 & 26.5 & 24.1 & 13.8 & 25.0 & 25.4 & 26.3 & 26.1\\
    C4AI-Command-R-v01 & \scorecolor{20.3} 29.7 & 30.0 & 29.7 & 29.6 & 30.0 & 29.9 & 29.6 & 29.5 & 29.4 & 29.8 & 29.7 & 29.8 & 29.2 & 29.7 & 29.9 & 29.9\\
    C4AI-Command-R+ & \scorecolor{20.0} 30.0 & 30.0 & 30.1 & 30.1 & 30.0 & 30.0 & 30.1 & 30.0 & 30.0 & 30.0 & 30.0 & 30.0 & 30.1 & 30.0 & 30.0 & 30.0\\
    \bottomrule
\end{tabular}
}
\caption{Comparing the average turn count broken down by scenario category and tool augmentations. Columns from left to right represent average turn count across all categories, then \textbf{S}ingle \textbf{T}ool \textbf{C}all, \textbf{M}ultiple \textbf{T}ool \textbf{C}all, \textbf{S}ingle \textbf{U}ser \textbf{T}urn, \textbf{M}ultiple \textbf{U}ser \textbf{T}urn, \textbf{S}tate \textbf{D}ependency, \textbf{C}anonicalization, \textbf{I}nsufficient \textbf{I}nformation, \textbf{0} \textbf{D}istraction \textbf{T}ools, \textbf{3} \textbf{D}istraction \textbf{T}ools, \textbf{10} \textbf{D}istraction \textbf{T}ools, \textbf{A}ll \textbf{T}ools, \textbf{T}ool \textbf{N}ame \textbf{S}crambled, \textbf{T}ool \textbf{D}escription \textbf{S}crambled, \textbf{A}rgument \textbf{D}escription \textbf{S}crambled and \textbf{A}rgument \textbf{T}ype \textbf{S}crambled. Note that turn count should not be interpreted in isolation, considering that a model could also be confidently wrong, finishing a task early without properly completing the user goal. As such, one should compare turn count between similarly similarity scored models, to showcase their difference in efficiency.} 
\label{tbl:avg_turn_count_per_category_all_models}
\end{table*}

\end{document}